\documentclass{article} 
\usepackage{iclr2027_conference,times}

\usepackage{amsmath,amsfonts,bm}

\def\eqref#1{equation~\ref{#1}}

\def\1{\bm{1}}

\DeclareMathAlphabet{\mathsfit}{\encodingdefault}{\sfdefault}{m}{sl}
\SetMathAlphabet{\mathsfit}{bold}{\encodingdefault}{\sfdefault}{bx}{n}

\usepackage{graphicx}
\usepackage{hyperref}
\usepackage{url}
\usepackage{amssymb}
\usepackage[ruled,vlined,linesnumbered]{algorithm2e}
\usepackage{booktabs}
\usepackage{multirow}
\usepackage{graphicx}
\usepackage{colortbl}
\usepackage{wrapfig}
\newcommand{\best}[1]{\textbf{#1}}
\newcommand{\second}[1]{\underline{#1}}
\SetKwInput{KwRequire}{Require}
\SetKwInput{KwOutput}{Output}
\DontPrintSemicolon
\title{TORE: Token-Wise Taylor Order Selection via Decomposed Residual Error Estimation for Diffusion Acceleration}

\author{Lei Tang, Baoquan Zhang, Yakun Zhang, Yunming Ye\thanks{Corresponding author.}\\
Harbin Institute of Technology, Shenzhen\\
{\footnotesize\texttt{25b351007@stu.hit.edu.cn, baoquanzhang@hit.edu.cn}}\\
{\footnotesize\texttt{zhangyk@hit.edu.cn, yeyunming@hit.edu.cn}}}

\iclrfinalcopy 
\begin{document}
\maketitle
\pagestyle{fancy}
\fancyhead[L]{Preprint.}
\begin{abstract}
Diffusion models excel at image and video generation, but iterative denoising incurs substantial inference latency. Recent token-adaptive acceleration methods select Taylor prediction orders for the full-model residual using shallow-layer proxy scores. However, candidate orders are evaluated on these proxy features but used to predict the full-model residual, creating a proxy--target mismatch: the proxy-preferred order may not minimize residual prediction error. To address this mismatch, we propose \textbf{TORE} (\textbf{T}aylor \textbf{O}rder selection via \textbf{R}esidual \textbf{E}rror estimation), which selects token-wise Taylor orders based on estimated prediction errors for the full-model residual. Yet direct error evaluation requires future ground-truth residuals, which are unavailable during order selection. TORE therefore estimates prediction error by analyzing how errors arise in Taylor extrapolation. Specifically, it decomposes prediction error into truncation error and finite-difference derivative error, then estimates both contributions from cached residuals to preselect token-wise orders for upcoming skip steps. Extensive experiments show that TORE achieves $4.31\times$ speedup on FLUX.1-dev and $4.18\times$ on Wan2.1 while maintaining high generation quality and fidelity.
\end{abstract}
\begin{figure*}[!ht]
    \centering
    \begin{minipage}{0.84\textwidth}
        \centering
        \includegraphics[width=\linewidth]{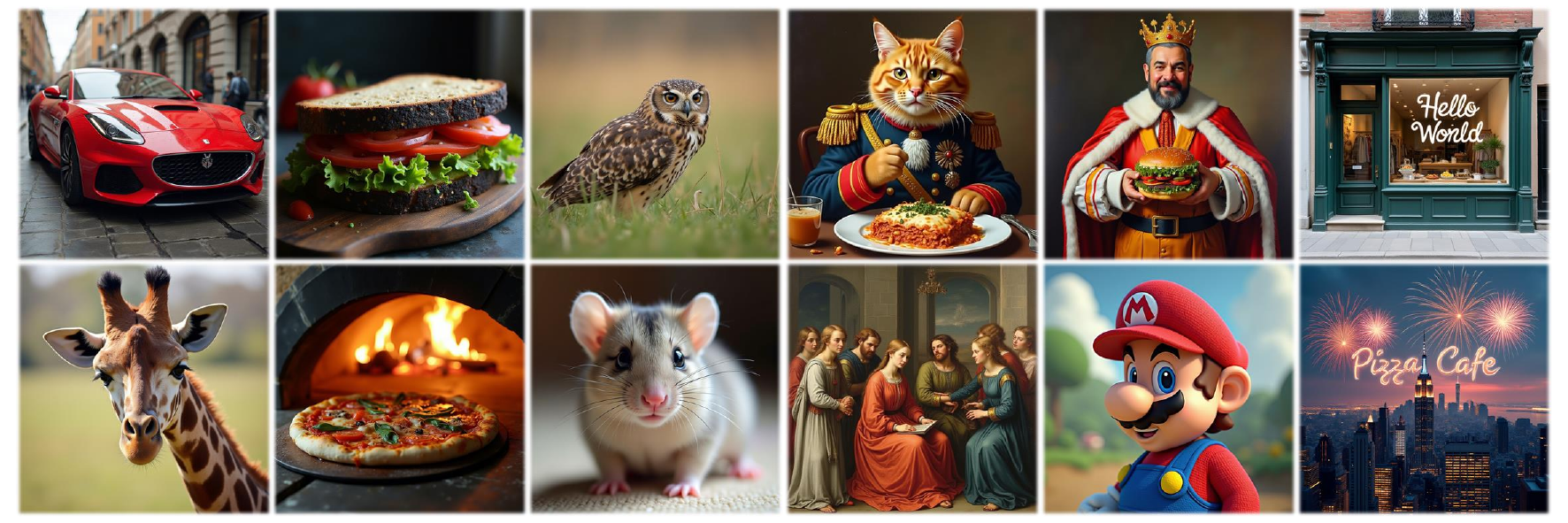}
        \caption{
        \textbf{Teaser of TORE.}
        Representative samples generated by TORE on FLUX.1-dev, achieving a \textbf{4.31$\times$} latency speedup while preserving high generation quality across diverse text prompts.
        }
        \label{fig:teaser}
    \end{minipage}
\end{figure*}

\section{Introduction}
Diffusion models have become a major class of generative models and have been widely applied to image and video generation \citep{Rombach2022LDM,Ho2022VideoDiffusion}. With the shift from U-Net to Transformer backbones, Diffusion Transformers (DiTs) achieve better generation quality but incur substantially higher computational costs~\citep{peebles2023scalable}. The iterative denoising process of diffusion models, combined with the quadratic complexity of self-attention, makes DiT inference very slow and costly. Therefore, accelerating DiT inference is important for practical deployment.

Training-free acceleration methods have recently emerged as an effective way to reduce the inference cost of diffusion models without additional training \citep{So2024FRDiff,Ma2024DeepCache,Zheng2025ClusterCache}. Existing caching approaches mainly exploit temporal redundancy across denoising steps, either by reusing cached features
\citep{Zou2025ToCa,Liu2025TeaCache,Kahatapitiya2025AdaCache} or by forecasting future features from historical observations \citep{Cui2026PredictToSkip,Han2026Spectrum}, with TaylorSeer using Taylor expansion for such prediction \citep{Liu2025TaylorSeer}. Building on TaylorSeer, TAP introduces token-wise adaptive Taylor order selection \citep{Zhu2026TAP}. It evaluates candidate orders on the modulated input to the first Transformer block, but applies the selected order to predict the full-model residual. This creates a proxy--target mismatch between order evaluation and residual prediction. This strategy implicitly assumes that candidate orders have consistent prediction-error rankings on shallow-layer features and on the full-model residual. However, this assumption does not always hold. A natural gap exists between the two quantities. Even strong feature correlation does not guarantee consistent prediction-error rankings. We verify this phenomenon in Sec.~\ref{sec:motivation}.
\begin{wrapfigure}{r}{0.5\textwidth}
    \centering
    \includegraphics[width=\linewidth]{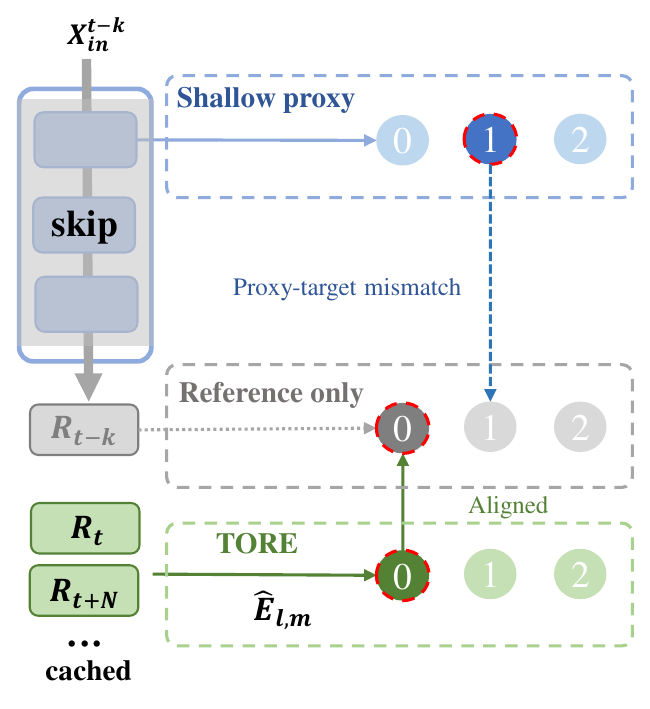}
    \caption{
    \textbf{Resolving the proxy--target mismatch.}
    Shallow-proxy scoring (blue) ranks Taylor orders by prediction
    errors on shallow features. TORE (green) instead ranks them
    by residual prediction errors estimated from cached residuals.
    }
    \label{fig:intro}
\end{wrapfigure}

To address the proxy--target mismatch, we propose \textbf{TORE} (\textbf{T}aylor \textbf{O}rder Selection via \textbf{R}esidual \textbf{E}rror Estimation). As illustrated in Fig.~\ref{fig:intro}, TORE bases order selection on the estimated error of the residual prediction target itself, rather than on a separate shallow proxy, thereby narrowing the proxy--target gap and better matching the oracle-optimal order. Yet measuring these errors directly requires ground-truth residuals at skipped steps, which are unavailable without full-model computation. Our key idea is to estimate prediction error by analyzing the approximations in the Taylor predictor, rather than by comparing predictions with ground truth. We estimate the prediction error from cached residuals and use it to select the lowest-error order for each token. The estimate captures two main error sources: \emph{Truncation Error}, caused by omitting the remainder of the Taylor expansion, and \emph{Finite-Difference Derivative Error}, introduced by approximating higher-order derivatives with finite differences. This decomposition also explains why Taylor order selection is important. For an ideal Taylor expansion, increasing the order reduces truncation error. In practice, however, higher-order derivatives are estimated using finite differences, and the resulting approximation error can accumulate as the order increases. Therefore, a higher order reduces truncation error but may increase finite-difference derivative error. The overall prediction error is determined by the balance between these two terms. The optimal order should thus minimize their combined error, and may vary across tokens. Besides, all error estimation and order selection are performed at full-computation steps, with the resulting token-wise orders reused for the subsequent skip steps without further estimation or selection.

Our main contributions are summarized as follows:

\textbf{TORE.}
We propose TORE, a training-free method for token-wise Taylor order selection in diffusion acceleration. TORE replaces shallow-proxy scores with estimates of each candidate order's prediction error on the full-model residual. This aligns order selection with the actual prediction target.

\textbf{Error Decomposition.}
We decompose Taylor prediction error into \emph{Truncation Error} and \emph{Finite-Difference Derivative Error}, and estimate both contributions from cached residuals. This provides a computable selection criterion without requiring future true residuals. TORE uses this criterion at each full-computation step to preselect orders for the next $N-1$ skip steps.

\textbf{Evaluation.}
Experiments on text-to-image, text-to-video, and class-to-image generation demonstrate favorable quality--latency trade-offs against four representative prediction-based baselines. On FLUX.1-dev, TORE achieves a $4.31\times$ speedup with an CLIP score of 28.020 while preserving high visual quality, as illustrated by the representative samples in Fig.~\ref{fig:teaser}. Further analyses show that our error estimates effectively guide token-wise Taylor order selection.

\section{Related Work}
\textbf{Fast Sampling and Few-Step Generation.} A major direction for diffusion acceleration is to reduce the number of denoising steps. Training-free samplers achieve this through improved numerical schemes \citep{Song2021DDIM,Liu2022PNDM,Lu2022DPMSolver,Zhao2023UniPC,Zhang2023DEIS}, whereas training-based approaches enable one- or few-step generation through distillation, consistency learning, or distribution matching \citep{Salimans2022ProgressiveDistillation,Song2023ConsistencyModels,Luo2023LatentConsistency,Yin2024DMD}. Rather than reducing the number of denoising updates, we retain the sampling schedule and replace selected full-network evaluations with residual forecasts whose Taylor orders are chosen per token according to estimated prediction error.

\textbf{Reuse-Based Caching Acceleration.}
Another direction exploits temporal redundancy to avoid repeated computation across denoising steps. This motivates reusing encoder or block features in U-Nets \citep{Li2024FasterDiffusion,Wimbauer2024BlockCaching} and cached representations in DiTs \citep{Chen2024DeltaDiT,Selvaraju2024FORA}. To make reuse more selective, layer-selection and attention-specific policies determine which representations to cache and when to refresh them \citep{Ma2024LearningToCache,Zhao2025PAB}. These policies refine where and when reuse occurs, but direct reuse remains a zero-order temporal approximation. We instead treat residual reuse as a candidate alongside higher-order forecasts, selecting the order with the lowest estimated residual prediction error for each token.

\textbf{Forecasting-Based Caching Acceleration.} Feature forecasting models temporal evolution rather than directly reusing cached representations. TaylorSeer uses Taylor expansion \citep{Liu2025TaylorSeer}, while alternative formulations include scaled Hermite polynomials in HiCache \citep{feng2026hicache}, rational approximation in TC-Pad\'e \citep{He2026TCPade}, and learned linear predictors \citep{Shen2026L2P}. Beyond predictor design, Confidence-Gated Taylor uses shallow-layer errors for reliability-based fallback \citep{Guan2025ConfidenceGatedTaylor}, while TAP uses shallow proxies for token-wise order--horizon selection \citep{Zhu2026TAP}. However, shallow-proxy errors need not preserve the ranking of candidate predictors at the full-model residual target. We instead construct computable token-wise estimates of residual prediction error by accounting for both Taylor truncation and derivative approximation errors. These estimates guide order selection from cached residuals, without requiring future ground-truth residuals or shallow-layer probes at skip steps.

\section{Method}
\subsection{Preliminaries}
\textbf{Diffusion Model.}
Diffusion models generate samples through an iterative denoising process, where the backbone is repeatedly evaluated over a sequence of timesteps. At timestep $t$, we denote the backbone input as $\mathbf{x}_{\mathrm{in}}\left(t\right)$ and its output as
$
\mathbf{x}_{\mathrm{out}}\left(t\right)
=
\Phi_{\theta}(\mathbf{x}_{\mathrm{in}}\left(t\right), t).
$
We define the \emph{full-model residual} as
\begin{equation}
\label{eq:full_residual}
\mathbf{R}\left(t\right)
=
\mathbf{x}_{\mathrm{out}}\left(t\right)
-
\mathbf{x}_{\mathrm{in}}\left(t\right),
\end{equation}
such that
$\mathbf{x}_{\mathrm{out}}\left(t\right)
=
\mathbf{x}_{\mathrm{in}}\left(t\right)
+
\mathbf{R}\left(t\right)$.
Instead of repeatedly evaluating the entire backbone at every timestep, we exploit the temporal redundancy across neighboring denoising steps and forecast the evolution of $\mathbf{R}_t$ from historical full-computation steps.

\textbf{Taylor-based forecasting.}
Taylor-based forecasting methods exploit the local approximation property of the Taylor expansion to extrapolate features at nearby timesteps. For computational efficiency, these methods typically truncate the Taylor series at a finite order and approximate the required high-order derivatives 
using finite differences. The resulting $m$-th-order predictor can be written as
\begin{equation}
\label{eq:taylor_prediction}
\widehat{\mathbf{R}}\left(t-k\right)
=
\mathbf{R}\left(t\right)
+
\sum_{i=1}^{m}
\frac{(-k)^i}{i!N^i}
\Delta^i \mathbf{R}\left(t\right),
\end{equation}
where $\mathbf{R}\left(t\right)$ denotes the residual at the reference timestep, $m$ is the Taylor expansion order, $k$ denotes the forecasting distance, and $N$ is the interval used to construct the finite differences. $\Delta^i \mathbf{R}\left(t\right)$ denotes the $i$-th-order finite difference of the residual. However, this formulation uses the same Taylor order $m$ for all tokens. In practice, tokens can evolve differently across denoising steps, and different prediction orders capture different patterns of temporal change. A single order may therefore not be optimal for every token. Selecting a suitable order for each token is thus important for improving Taylor-based acceleration.
\subsection{Motivation Analysis}
\label{sec:motivation}
\begin{figure}[t]
    \centering
    \includegraphics[width=0.75\linewidth]{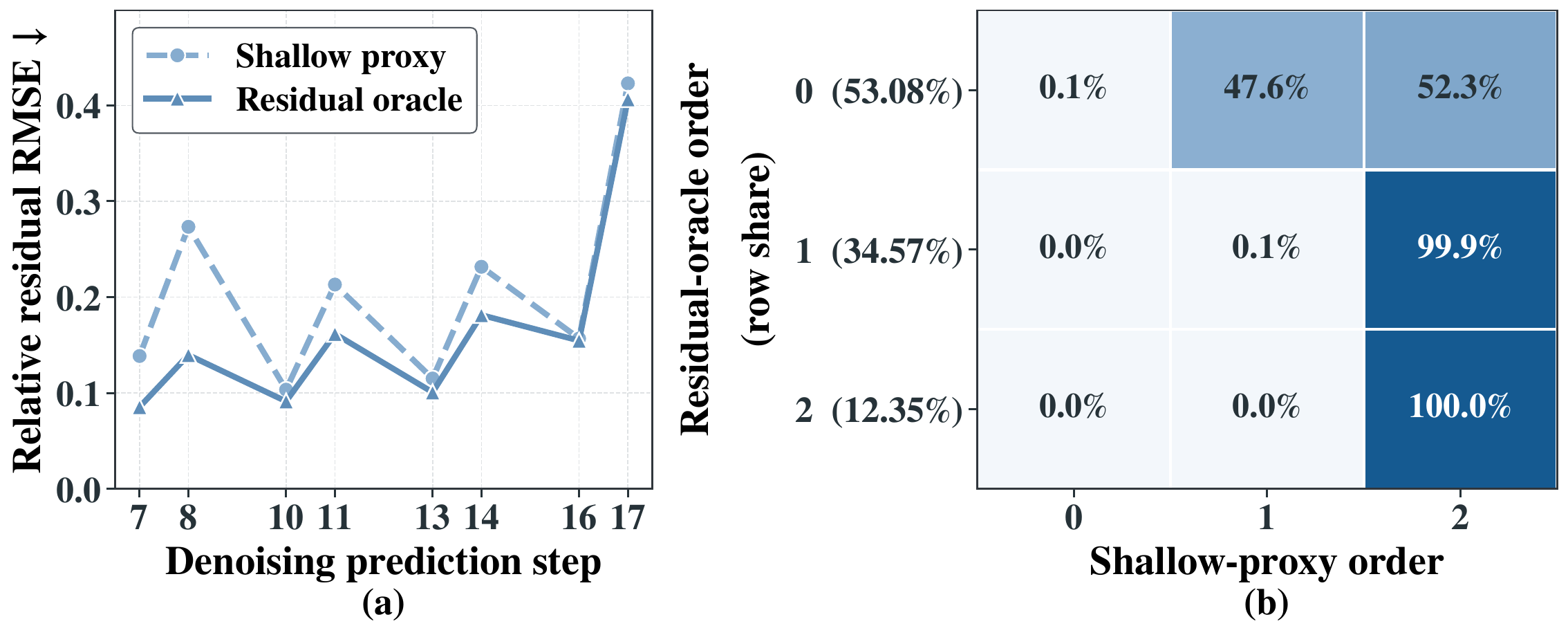}
    \caption{ 
    \textbf{Limitations of shallow-proxy order selection.}
    (a) Residual prediction error. (b) Shallow-proxy order distributions conditioned on the residual-oracle order; parentheses indicate each row’s sample share.
    }
    \label{fig:motivation}
\end{figure}
Shallow-proxy selection assumes that the order preferred on shallow features is also suitable for predicting the full-model residual. However, the two targets need not favor the same Taylor order.
We examine this issue using an 18-step sampling schedule, $N=3$, $M=2$, and a resolution of $256\times256$. At each scheduled skip step, we obtain a reference residual through full-network computation and measure the error of the proxy-selected prediction. For comparison, a \emph{residual oracle} uses this reference to select the lowest-MSE candidate order for each token. Figure~\ref{fig:motivation}(a) shows that shallow-proxy selection incurs higher normalized RMSE than the oracle at several steps. Since both use the same candidate predictions, this gap indicates
that shallow-proxy scores do not always identify the best order for the target residual.

We further compare the orders selected by the shallow proxy and the residual oracle in Fig.~\ref{fig:motivation}(b). Rows correspond to oracle orders and columns to proxy orders. Each row shows the distribution of proxy selections among tokens with the corresponding oracle order. The percentages in parentheses indicate each row's share of all evaluated tokens. Diagonal entries therefore show the percentage of tokens in each row for which the proxy and oracle select the same order. The shallow proxy strongly favors higher orders. It matches the oracle on $100\%$ of tokens with oracle order 2, but on only $0.1\%$ of tokens in each of the oracle-order-0 and oracle-order-1 groups. These two groups together account for $87.65\%$ of all evaluated tokens. This discrepancy shows that the order preferred by the shallow proxy is not necessarily optimal for predicting the full-model residual and it also motivates selecting orders based on estimated residual prediction error rather than shallow-proxy scores.
\subsection{TORE: taylor order selection via residual error estimation}
\subsubsection{Error-Guided Residual Forecasting}
\begin{figure}[t]
    \centering
    \includegraphics[width=\linewidth]{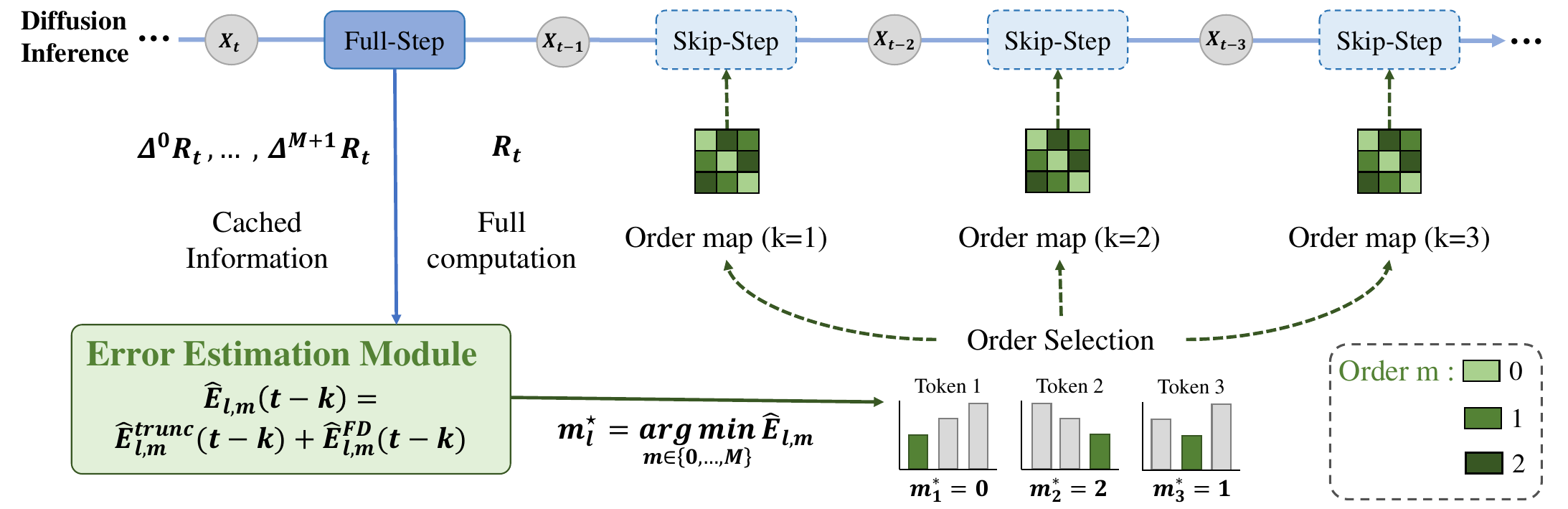}
    \caption{
        \textbf{Overview of TORE.}
        Full-steps preselect Taylor orders for each token
        at each of the next $N-1$ skip steps.
        Skip steps use the stored orders for residual forecasting,
        without further error estimation or order selection.
    }
    \label{fig:framework}
\end{figure}

We formulates Taylor order selection as residual-error minimization and construct a computable selection criterion by estimating the approximation errors in Taylor forecasting. As illustrated in Fig.~\ref{fig:framework}, this criterion determines which order to use for each token and prediction horizon, with order evaluation and forecasting sharing the full-model residual as their target.

Let $t$ denote the current full-computation step, $N$ the interval between full steps, and $k \in \{1,\ldots,N-1\}$ a prediction horizon. For token $l$, let $\widehat{R}_{l,m}(t-k)$ denote the order-$m$ forecast anchored at $t$, where $m \in \{0,\ldots,M\}$. Our objective is to minimize $\|R_l(t-k)-\widehat{R}_{l,m}(t-k)\|_2$. Since obtaining the target residual requires full-model computation, we instead evaluate candidate orders using the cached residual
finite differences $\{\Delta_N^i R(t)\}_{i=0}^{M+1}$, with $\Delta_N^0 R(t)=R(t)$.

The \emph{Error Estimation Module} assigns each candidate the score
\begin{equation}
\widehat{E}_{l,m}(t-k)
=
\widehat{E}^{\mathrm{trunc}}_{l,m}(t-k)
+
\widehat{E}^{\mathrm{FD}}_{l,m}(t-k).
\label{eq:tore_error_score}
\end{equation}
The two terms estimate the contributions from omitted Taylor terms and finite-difference approximations of the retained derivatives, respectively. Their combination provides a criterion for comparing candidate orders without requiring an exact evaluation of the future residual error. Minimizing this score yields the token- and horizon-specific order map:
\begin{equation}
\mathcal{O}[k,l]
=
\widehat{m}_l(t-k)
=
\arg\min_{m\in\{0,\ldots,M\}}
\widehat{E}_{l,m}(t-k).
\label{eq:tore_order_map}
\end{equation}
In Fig.~\ref{fig:framework}, the bars compare candidate-order scores for individual tokens, while each grid records the selected orders for one prediction horizon. Order $0$ corresponds to direct residual reuse, whereas higher orders incorporate additional temporal variation. TORE precomputes these maps at the full-computation step $t$. At each skip step $t-k$, TORE performs token-wise Taylor forecasting using the corresponding precomputed order map, then adds the predicted residual to the network input to obtain the output. No further error estimation or order selection is required. The next subsection derives the two error estimates from cached residual differences under a local smoothness assumption. Appendix~A provides the complete inference procedure.
\subsubsection{Decomposed Residual Error Estimation}
We now derive the two components of ~\eqref{eq:tore_error_score}. Assuming that the residual trajectory is locally smooth around $t$, the order-$m$ Taylor expansion of the target residual is
\begin{equation}
R_l(t-k)
=
R_l(t)
+
\sum_{i=1}^{m}
\frac{(-k)^i}{i!}R_l^{(i)}(t)
+
\mathcal{R}_{l,m+1},
\end{equation}
where $\mathcal{R}_{l,m+1}$ denotes the Taylor remainder. In practical forecasting, the derivatives are approximated using finite differences computed from cached residuals. Subtracting the predictor in ~\eqref{eq:taylor_prediction} from the exact expansion gives
\begin{equation}
e_{l,m}=
\mathcal{R}_{l,m+1}
+
\sum_{i=1}^{m}
\frac{(-k)^i}{i!}
\left(
R_l^{(i)}(t)
-
\frac{\Delta_N^i R_l(t)}{N^i}
\right),
\label{eq:error_decomposition}
\end{equation}
~\eqref{eq:error_decomposition} gives an exact decomposition of the prediction error. The first term is the truncation error introduced by omitting the Taylor remainder, while the second term is the error caused by replacing
the exact derivatives with finite differences. Neither component can be evaluated exactly at a forecasted step because the remainder and the exact derivatives are unavailable. We therefore construct local estimates from the finite differences of previously computed residuals. These estimates are used to compare candidate orders rather than to recover the exact error magnitude.

Then, we define
\begin{equation}
d_{l,i}\left(t\right)
=
\left\|
\Delta_N^i R_l(t)
\right\|_2,
\label{eq:finite_difference_norm}
\end{equation}
where $\Delta_N^i R_l(t)$ denotes the $i$-th-order finite difference of the 
residual $R_l(t)$ with interval $N$, and $\|\cdot\|_2$ denotes the $L_2$ norm.

\textbf{Proposition 1 (Truncation Error Estimate).}
Under the local smoothness assumption, we approximate the Taylor remainder using the first omitted term and estimate its derivative from the corresponding finite difference. This gives the local truncation error estimate
\begin{equation}
\widehat{E}_{l,m}^{\mathrm{trunc}}(t-k)
=
\frac{k^{m+1}}
{(m+1)!N^{m+1}}
d_{l,m+1}(t).
\label{eq:truncation_error}
\end{equation}

\textbf{Proposition 2 (Finite Difference Approximation Error Estimate).}
The local expansions of the $i$-th and $(i+1)$-st finite differences give
\begin{equation}
\label{eq:fd_error_expansion}
R_l^{(i)}(t)
-
\frac{\Delta_N^i R_l(t)}{N^i}
=
-\frac{i}{2N^i}
\Delta_N^{i+1}R_l(t)
+
\rho_{l,i}(t),
\end{equation}
where $\rho_{l,i}(t)$ collects the remaining terms in the local expansion. Thus, $\Delta_N^{i+1}R_l(t)$ is not treated as an exact substitute for the unknown derivative. Instead, it provides the first observable correction to the $i$-th finite difference approximation. After applying the corresponding Taylor coefficients, we sum the magnitudes of the retained correction terms to obtain
\begin{equation}
    \widehat{E}_{l,m}^{\mathrm{FD}}(t-k)
    =
    \frac{1}{2}
    \sum_{i=1}^{m}
    \frac{k^i}{(i-1)!N^i}
    d_{l,i+1}(t).
\label{eq:fd_error}
\end{equation}

Taking the norm of each contribution before summation produces a nonnegative
score and avoids cancellation between derivative approximation terms. The
detailed derivation of both estimators is provided in the Appendix \ref{app:error_derivation}.

\section{Experiments}
\begin{table*}[t]
    \centering
    \caption{
        Text-to-image generation results on DrawBench using FLUX.1-dev.
        The best and second-best quality results within each comparison group
        are marked in \textbf{bold} and \underline{underline}, respectively.
    }
    \label{tab:text2image_results}

    \setlength{\tabcolsep}{4.2pt}
    \renewcommand{\arraystretch}{1.12}

    \resizebox{\textwidth}{!}{%
    \begin{tabular}{@{}lccccccccc@{}}
        \toprule
        \multirow{2}{*}{\textbf{Method}}
        & \multicolumn{4}{c}{\textbf{Efficiency}}
        & \multicolumn{5}{c}{\textbf{Generation Quality}} \\
        \cmidrule(lr){2-5}
        \cmidrule(lr){6-10}
        &
        \shortstack{\textbf{Latency}\\\textbf{(s)} $\downarrow$}
        &
        \shortstack{\textbf{Latency}\\\textbf{Speedup} $\uparrow$}
        &
        \shortstack{\textbf{FLOPs(T)}\\$\downarrow$}
        &
        \shortstack{\textbf{FLOPs}\\\textbf{Speedup} $\uparrow$}
        &
        \shortstack{\textbf{Image}\\\textbf{Reward} $\uparrow$}
        &
        \shortstack{\textbf{CLIP}\\\textbf{Score} $\uparrow$}
        &
        \textbf{PSNR} $\uparrow$
        &
        \textbf{SSIM} $\uparrow$
        &
        \textbf{LPIPS} $\downarrow$ \\
        \midrule

        Full
        & 16.27
        & 1.00$\times$
        & 3719.23
        & 1.00$\times$
        & 1.0414
        & 28.010
        & --
        & --
        & -- \\

        60\% steps
        & 9.77
        & 1.66$\times$
        & 2232.00
        & 1.66$\times$
        & 1.0069 
        & 27.440
        & 19.060
        & 0.7401
        & 0.2477 \\
        
        50\% steps
        & 8.11
        & 2.00$\times$
        & 1860.00
        & 1.99$\times$
        & 1.0126
        & 27.607
        & 17.408
        & 0.6947
        & 0.3102 \\
        
        40\% steps
        & 6.48
        & 2.51$\times$
        & 1488.00
        & 2.49$\times$
        & 0.9418
        & 27.505
        & 16.363
        & 0.6604
        & 0.3599 \\

        \midrule

        TaylorSeer (N=2,O=1)
        & 8.84
        & 1.84$\times$
        & 1935.21
        & 1.92$\times$
        & \best{1.0494}
        & \second{27.873}
        & 22.142
        & 0.8043
        & 0.1575 \\

        HiCache (N=2,O=2)
        & 10.79
        & 1.50$\times$
        & 1935.21
        & 1.92$\times$
        & 0.9913
        & 27.668
        & \second{23.953}
        & 0.8109
        & \second{0.1394} \\
        
        TC-Padé (N=3,$\theta$=1.0)
        & 7.16
        & 2.27$\times$
        & 1562.54
        & 2.38$\times$
        & 0.9968
        & 27.608
        & 23.035
        & \best{0.8384}
        & 0.1546 \\

        TAP (N=2)
        & 8.47
        & 1.92$\times$
        & 1937.09
        & 1.92$\times$
        & 1.0224
        & 27.826
        & 21.368
        & 0.7928
        & 0.1817 \\
        
        \textbf{TORE (N=3,M=4)}
        & 6.46
        & 2.51$\times$
        & 1487.79
        & 2.49$\times$
        & \second{1.0462}
        & \best{28.043}
        & \best{24.450}
        & \second{0.8291}
        & \best{0.1188} \\

        \midrule

        TaylorSeer (N=3,O=2)
        & 6.75
        & 2.41$\times$
        & 1414.01
        & 2.63$\times$
        & 1.0266
        & 27.919
        & 19.297
        & 0.7345
        & 0.2391 \\

        HiCache (N=3,O=2)
        & 6.58
        & 2.47$\times$
        & 1340.18
        & 2.77$\times$
        & \second{1.0275}
        & \second{27.945}
        & 19.554
        & 0.7384
        & 0.2319 \\

        TC-Padé (N=6,$\theta$=1.0)
        & 5.41
        & 3.00$\times$
        & 1116.63
        & 3.33$\times$
        & 0.9808
        & 27.509
        & \second{22.034}
        & \second{0.8043}
        & \second{0.2067} \\

        TAP (N=3)
        & 5.74
        & 2.83$\times$
        & 1339.27
        & 2.77$\times$
        & 0.9781
        & 27.559
        & 18.683
        & 0.7281
        & 0.2588 \\

        \textbf{TORE (N=5,M=6)}
        & 5.17
        & 3.14$\times$
        & 1190.53
        & 3.12$\times$
        & \best{1.0364}
        & \best{28.011}
        & \best{24.387}
        & \best{0.8210}
        & \best{0.1302} \\

        \midrule

        TaylorSeer (N=5,O=2)
        & 4.83
        & 3.36$\times$
        & 968.50
        & 3.84$\times$
        & \best{1.0528}
        & 27.819
        & 16.657
        & 0.6546
        & 0.3600 \\

        HiCache (N=4,O=2)
        & 5.25
        & 3.09$\times$
        & 1042.68
        & 3.56$\times$
        & 1.0252
        & \second{27.879}
        & 18.023
        & 0.6893
        & 0.2942 \\

        TC-Padé (N=8,$\theta$=1.4)
        & 4.13
        & 3.93$\times$
        & 893.47
        & 4.16$\times$
        & 0.9469
        & 27.430
        & \second{20.644}
        & \best{0.7679}
        & \second{0.2728} \\

        TAP (N=5)
        & 4.11
        & 3.95$\times$
        & 967.28
        & 3.84$\times$
        & 0.9431
        & 27.585
        & 15.583
        & 0.6410
        & 0.3998 \\
        
        \textbf{TORE (N=8,M=5)}
        & 3.77
        & 4.31$\times$
        & 892.94
        & 4.16$\times$
        & \second{1.0308}
        & \best{28.020}
        & \best{21.699}
        & \second{0.7582}
        & \best{0.2126} \\

        \bottomrule
    \end{tabular}%
    }
\end{table*}
\subsection{Experiment Settings}
We conduct comprehensive experiments on three representative generative models 
across different generation tasks. For text-to-image generation, we evaluate 
FLUX.1-dev \citep{labs2024flux} using the 200 prompts from DrawBench. 
For text-to-video generation, we evaluate Wan2.1 \citep{wan2025wan} following 
the standard VBench evaluation protocol. For class-to-image generation, 
we evaluate DiT-XL/2 \citep{peebles2023scalable} on ImageNet. 
These settings follow widely adopted evaluation protocols, enabling fair 
comparisons with existing acceleration methods.
\subsection{Experiment Results}
\begin{figure}[t]
    \centering
    \includegraphics[width=0.8\linewidth]{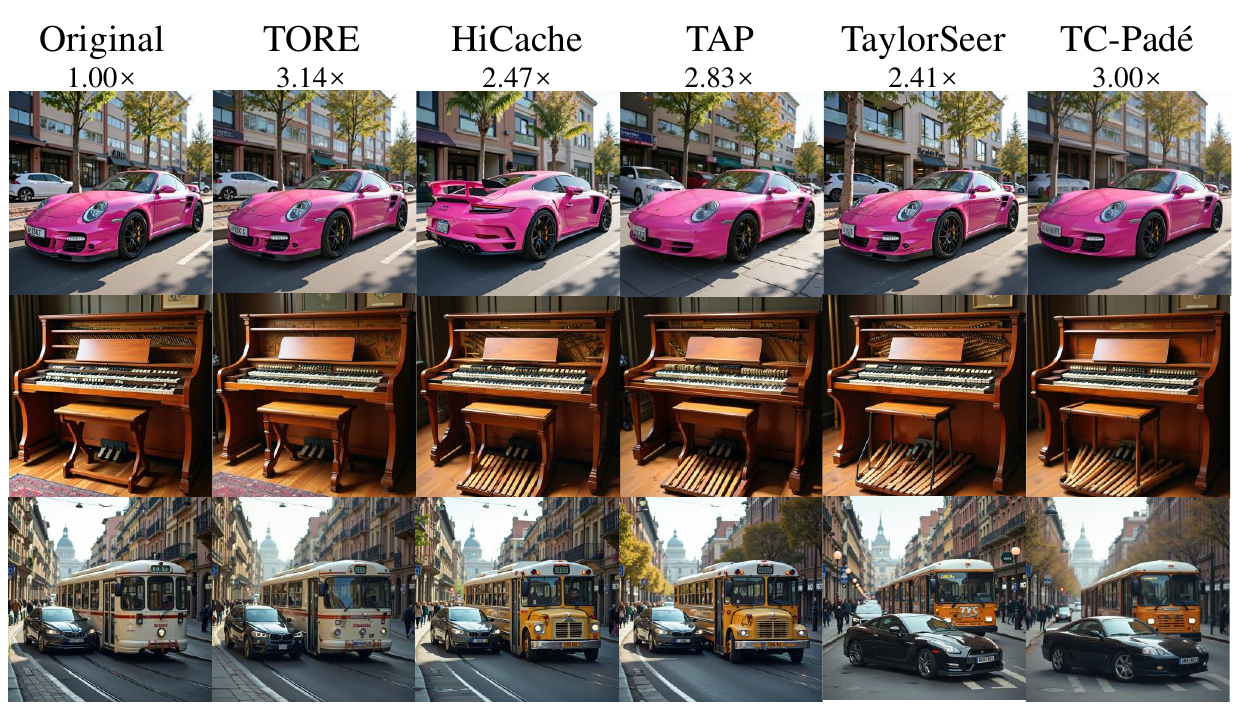}
    \caption{Qualitative comparison of different acceleration methods on the
text-to-image generation task across diverse prompts.
All methods use the same prompts and generation settings.}
    \label{fig:qualitative}
\end{figure}

\begin{figure}[!ht]
    \centering
    \includegraphics[width=\linewidth]{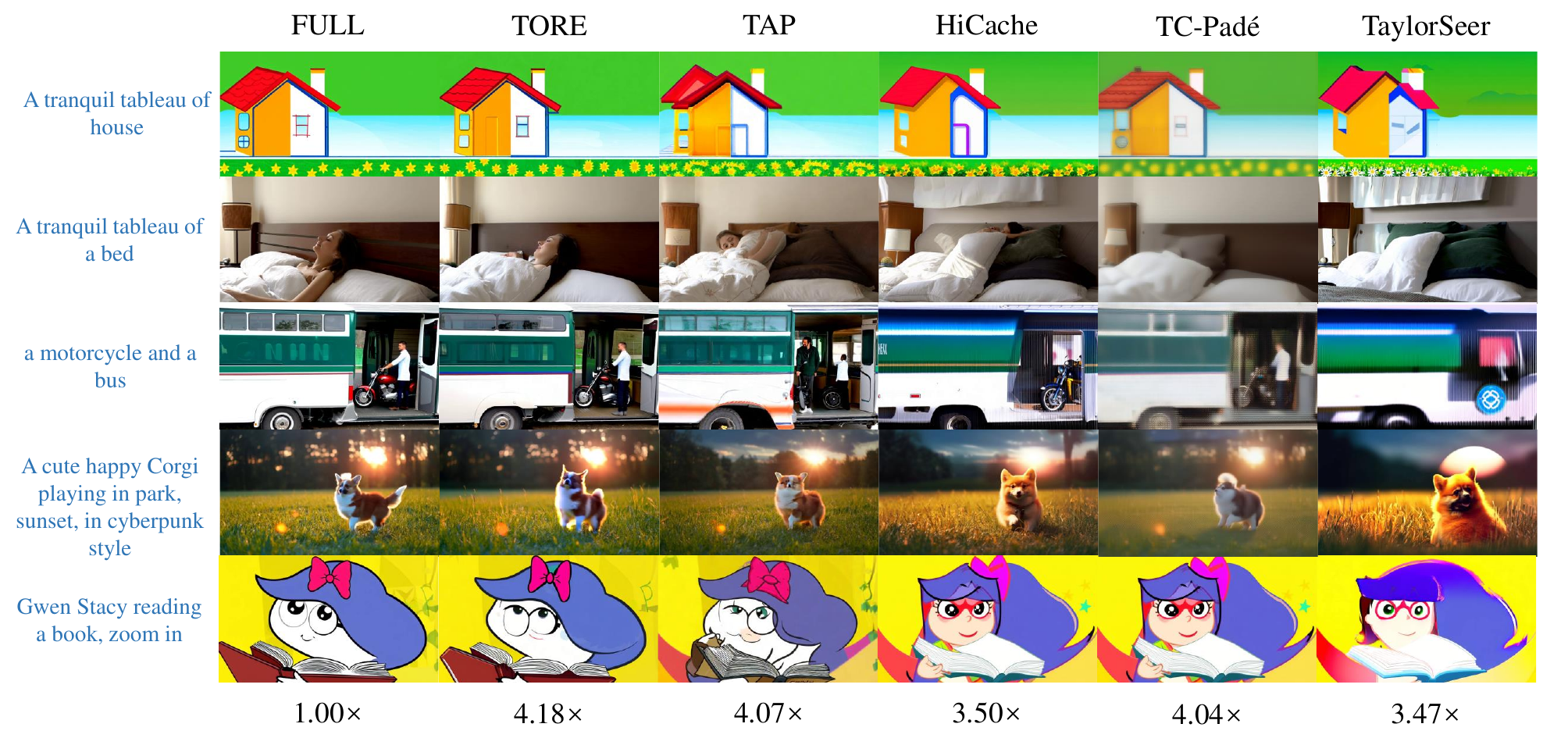}
    \caption{
    \textbf{Qualitative comparison on text-to-video generation.}
    Representative frames generated by different acceleration methods under 
    comparable speedup ratios. TORE achieves a $4.18\times$ latency speedup 
    while preserving high visual consistency with the full-inference baseline 
    across diverse prompts.
    }
    \label{fig:video_qualitative}
\end{figure}

\textbf{Text-to-Image Generation}. We evaluate TORE on FLUX.1-dev against TaylorSeer, TAP, HiCache, and TC-Pad\'{e} at multiple forecasting intervals. As shown in Table~\ref{tab:text2image_results}, TORE achieves favorable quality--latency trade-offs. In the second comparison group, TORE achieves a $3.14\times$ speedup and the best results on all five quality metrics, including an ImageReward score of 1.0364. Compared with TC-Pad\'{e} at $3.00\times$ speedup, TORE improves PSNR by $2.35$ dB and reduces LPIPS by $37.0\%$. At the more aggressive setting, TORE achieves a $4.31\times$ speedup. Compared with TC-Pad\'{e} in this group, it reduces latency by $8.7\%$ and LPIPS by $22.1\%$ at a comparable FLOP count. Figure~\ref{fig:qualitative} compares outputs generated with the same prompts. In these examples, TORE better preserves the scene layout and visual details of the full-inference baseline despite its higher speedup.
\begin{table}[t]
    \centering
    \caption{
        Comparison on text-to-video generation for Wan2.1 on VBench.
        The best results among accelerated methods within each comparison
        group are shown in \textbf{bold}.
    }
    \label{tab:text2video}

    \setlength{\tabcolsep}{3.3pt}
    \renewcommand{\arraystretch}{1.12}

    \resizebox{0.85\linewidth}{!}{%
    \begin{tabular}{@{}lccccccc@{}}
        \toprule
        \multirow{2}{*}{\textbf{Method}}
        & \multicolumn{4}{c}{\textbf{Efficiency}}
        & \multicolumn{3}{c}{\textbf{VBench} $\uparrow$} \\
        \cmidrule(lr){2-5}
        \cmidrule(lr){6-8}
        &
        \shortstack{\textbf{Latency}\\(s) $\downarrow$}
        &
        \shortstack{\textbf{Latency}\\\textbf{Speedup} $\uparrow$}
        &
        \shortstack{\textbf{FLOPs(T)}\\$\downarrow$}
        &
        \shortstack{\textbf{FLOPs}\\\textbf{Speedup} $\uparrow$}
        &
        \shortstack{\textbf{Overall}\\\textbf{Score}}
        &
        \shortstack{\textbf{Quality}\\\textbf{Score}}
        &
        \shortstack{\textbf{Semantic}\\\textbf{Score}} \\
        \midrule

        Full
        & 131.51
        & 1.00$\times$
        & 28303.52
        & 1.00$\times$
        & 80.72
        & 84.41
        & 65.98 \\

        \midrule
        TaylorSeer (N=4,O=1)
        & 42.22
        & 3.11$\times$
        & 7927.24
        & 3.57$\times$
        & 77.24
        & 82.18
        & 57.48 \\
        
        HiCache (N=4,O=1)
        & 42.22
        & 3.11$\times$
        & 7927.24
        & 3.57$\times$
        & 79.03
        & 83.19
        & 62.41 \\
        
        TC-Padé (N=6,$\theta$=1.0)
        & 41.09
        & 3.20$\times$
        & 8492.35
        & 3.33$\times$
        & 79.83
        & 83.66
        & 64.53 \\

        TAP (N=4)
        & 37.68
        & 3.49$\times$
        & 7928.15
        & 3.57$\times$
        & 79.86
        & 83.47
        & 65.43 \\


        \textbf{TORE (N=6,M=5)}
        & 36.96
        & 3.56$\times$
        & 7925.99
        & 3.57$\times$
        & \textbf{80.47}
        & \textbf{84.13}
        & \textbf{65.82} \\

        \midrule

        TaylorSeer (N=5,O=1)
        & 37.82
        & 3.47$\times$
        & 6795.20
        & 4.16$\times$
        & 74.46
        & 80.09
        & 51.96 \\

        HiCache (N=5,O=1)
        & 37.57
        & 3.50$\times$
        & 6795.20
        & 4.16$\times$
        & 77.86
        & 82.51
        & 59.27 \\

        TC-Padé (N=8,$\theta$=1.4)
        & 32.55
        & 4.04$\times$
        & 6794.23
        & 4.16$\times$
        & 75.29
        & 79.69
        & 57.71 \\        

        TAP (N=5)
        & 32.31
        & 4.07$\times$
        & 6795.56
        & 4.16$\times$
        & 79.71
        & 83.14
        & \textbf{65.99} \\
        
        \textbf{TORE (N=7,M=4)}
        & 31.46
        & 4.18$\times$
        & 6793.88
        & 4.16$\times$
        & \textbf{80.22}
        & \textbf{83.87}
        & 65.64 \\


        \bottomrule
    \end{tabular}%
    }
\end{table}
\textbf{Text-to-Video Generation.} We evaluate TORE on Wan2.1 using VBench. As shown in Table~\ref{tab:text2video}, TORE achieves the lowest latency and the highest overall and quality scores among accelerated methods in both comparison groups. At a $3.56\times$ speedup, TORE obtains an overall score of 80.47, only 0.25 points below the full-inference baseline. At a $4.18\times$ speedup, it achieves an overall score of 80.22, with a drop of 0.50 points from the baseline. Compared with TAP in this group, TORE improves the overall and quality scores by 0.51 and 0.73 points, respectively, while maintaining slightly lower latency. Figure~\ref{fig:video_qualitative} compares generated videos at similar speedups. In these examples, TORE better preserves the semantic content and visual appearance of the full-inference baseline.

\textbf{Class-to-Image Generation.}
We evaluate TORE on ImageNet using DiT-XL/2. As shown in Table~\ref{tab:class_generation}, TORE achieves the lowest FID and sFID and the highest latency speedup among forecasting methods in both comparison groups. At a $1.56\times$ speedup, TORE achieves an FID of 2.47, only 0.06 above the full-inference baseline, and matches its sFID of 4.39. At a $1.90\times$ speedup, TORE obtains an FID of 2.67
and an sFID of 4.40. Compared with using $50\%$ of the original sampling steps, TORE reduces FID by $15.2\%$ and sFID by $7.8\%$ at the same reported latency of 0.22 seconds.
\begin{table}[t]
    \centering
    \caption{
        Comparison for class-to-image generation on ImageNet with DiT-XL/2.
        The best results among accelerated methods within each comparison
        group are shown in \textbf{bold}.
    }
    \label{tab:class_generation}

    \setlength{\tabcolsep}{3.2pt}
    \renewcommand{\arraystretch}{1.12}

    \resizebox{0.8\linewidth}{!}{%
    \begin{tabular}{@{}lccccccc@{}}
        \toprule
        \multirow{2}{*}{\textbf{Method}}
        & \multicolumn{4}{c}{\textbf{Efficiency}}
        & \multicolumn{3}{c}{\textbf{Generation Quality}} \\
        \cmidrule(lr){2-5}
        \cmidrule(lr){6-8}
        &
        \shortstack{\textbf{Latency}\\(s) $\downarrow$}
        &
        \shortstack{\textbf{Latency}\\\textbf{Speedup} $\uparrow$}
        &
        \shortstack{\textbf{FLOPs(T)}\\$\downarrow$}
        &
        \shortstack{\textbf{FLOPs}\\\textbf{Speedup} $\uparrow$}
        &
        \textbf{FID} $\downarrow$
        &
        \textbf{sFID} $\downarrow$
        &
        \shortstack{\textbf{Inception}\\\textbf{Score} $\uparrow$} \\
        \midrule

        Full
        & 0.42
        & 1.00$\times$
        & 23.73
        & 1.00$\times$
        & 2.41
        & 4.39
        & 236.31 \\
        
        60\% steps
        & 0.26
        & 1.61$\times$
        & 14.23
        & 1.66$\times$
        & 2.81
        & 4.49
        & 228.78 \\
        
        50\% steps
        & 0.22
        & 1.90$\times$
        & 11.86
        & 2.00$\times$
        & 3.15
        & 4.77
        & 227.45 \\
        \midrule

        TaylorSeer (N=3,O=2)
        & 0.35
        & 1.18$\times$
        & 8.59
        & 2.76$\times$
        & 2.56
        & 4.66
        & 233.72 \\

        HiCache (N=2,O=1)
        & 0.33
        & 1.25$\times$
        & 12.42
        & 1.91$\times$
        & 2.64
        & 4.55
        & 234.06 \\

        TC-Padé (N=2,$\theta$=0.8)
        & 0.29
        & 1.42$\times$
        & 13.33
        & 1.78$\times$
        & 3.53
        & 5.27
        & 221.81 \\
        
        TAP (N=2)
        & 0.29
        & 1.45$\times$
        & 13.33
        & 1.78$\times$
        & 2.48
        & 4.41
        & \textbf{234.26} \\
        
        \textbf{TORE (N=2,M=3)}
        & 0.26
        & 1.61$\times$
        & 12.82
        & 1.85$\times$
        & \textbf{2.47}
        & \textbf{4.39}
        & 232.68 \\

        \midrule

        TaylorSeer (N=5,O=2)
        & 0.31
        & 1.34$\times$
        & 5.27
        & 4.50$\times$
        & 2.91
        & 5.25
        & 227.11 \\

        HiCache (N=4,O=2)
        & 0.28
        & 1.49$\times$
        & 6.70
        & 3.54$\times$
        & 3.63
        & 6.30
        & 222.44 \\

        TC-Padé (N=3,$\theta$=1.0)
        & 0.25
        & 1.66$\times$
        & 9.97
        & 2.38$\times$
        & 5.40
        & 6.77
        & 199.91 \\

        TAP (N=4)
        & 0.24
        & 1.75$\times$
        & 8.09
        & 2.93$\times$
        & 2.88
        & 4.58
        & 224.56 \\

        \textbf{{TORE (N=3,M=2)}}
        & 0.22
        & 1.90$\times$
        & 9.02
        & 2.63$\times$
        & \textbf{2.67}
        & \textbf{4.40}
        & \textbf{229.29} \\

        \bottomrule
    \end{tabular}%
    }
\end{table}
\subsection{Ablation and Analysis}
\begin{figure}[!ht]
    \centering
    \includegraphics[width=0.8\linewidth]{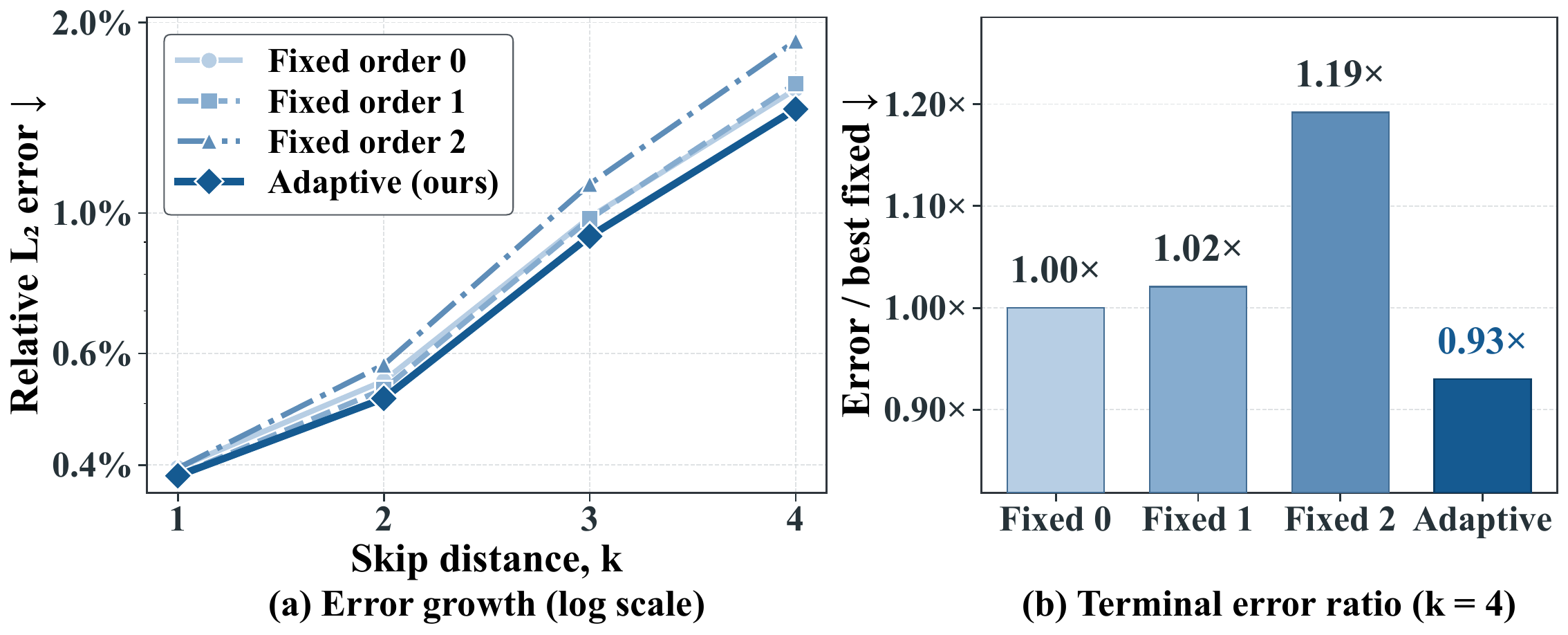}
    \caption{
    \textbf{Effectiveness of error-guided token-wise order selection.}
    \textbf{Left:} Relative $L_2$ residual prediction error versus skip distance $k$.
    \textbf{Right:} Normalized error at $k=4$, with the best fixed-order
    baseline (\texttt{fixed-0}) set to $1$.
    }
    \label{fig:order_selection}
\end{figure}

\textbf{Effect of Adaptive Order Selection.}
With a full-step interval of $N=5$ and maximum order $M=2$, we independently forecast the next four steps from the same full-computation anchor to exclude recursive error propagation. Full-model residuals at each target step serve as ground truth. We compare our adaptive strategy with \texttt{fixed-0}, \texttt{fixed-1}, and \texttt{fixed-2}, which apply a uniform Taylor order to all tokens. Our strategy outperforms all fixed-order baselines across forecasting distances, reducing the error at $k=4$ by $7\%$ relative to the best fixed order (Fig.~\ref{fig:order_selection}).

\textbf{Component Ablation.}
We conduct a $2\times2$ ablation to isolate the effects of full-model residual prediction and error-guided order selection.
As shown in Table~\ref{tab:component_ablation}, full-model residual prediction consistently reduces latency with nearly unchanged FLOPs, while error-guided order selection substantially improves reconstruction fidelity in terms of PSNR, SSIM, and LPIPS. Combining both components achieves a favorable efficiency-quality trade-off, reaching a $3.32\times$ latency speedup with the best reconstruction fidelity.
\begin{table}[!htbp]
    \centering
    \caption{
        Ablation study of full-model residual prediction and
        token-adaptive order selection.
        $\checkmark$ and $\times$ denote whether each component is enabled.
        Best results are shown in \textbf{bold}.
    }
    \label{tab:component_ablation}

    \small
    \setlength{\tabcolsep}{2.6pt}
    \renewcommand{\arraystretch}{1.15}

    \begin{tabular}{@{}ccccccccccc@{}}
        \toprule
        \multicolumn{2}{c}{\textbf{Components}}
        & \multicolumn{4}{c}{\textbf{Efficiency}}
        & \multicolumn{5}{c}{\textbf{Generation Quality}} \\
        \cmidrule(lr){1-2}
        \cmidrule(lr){3-6}
        \cmidrule(lr){7-11}

        \shortstack{\textbf{Residual}\\\textbf{Pred}}
        &
        \shortstack{\textbf{Token}\\\textbf{Adaptive}}
        &
        \shortstack{\textbf{Latency}\\(s) $\downarrow$}
        &
        \shortstack{\textbf{Speed}\\$\uparrow$}
        &
        \shortstack{\textbf{FLOPs(T)}\\$\downarrow$}
        &
        \shortstack{\textbf{Speed}\\$\uparrow$}
        &
        \shortstack{\textbf{Image}\\\textbf{Reward} $\uparrow$}
        &
        \shortstack{\textbf{CLIP}\\\textbf{Score} $\uparrow$}
        &
        \textbf{PSNR} $\uparrow$
        &
        \textbf{SSIM} $\uparrow$
        &
        \textbf{LPIPS} $\downarrow$ \\

        \midrule

        $\times$ & $\times$
        & 5.81
        & 2.80$\times$
        & 1042.88
        & 3.56$\times$
        & \textbf{1.0477}
        & 27.810
        & 17.646
        & 0.6821
        & 0.3087 \\

        $\checkmark$ & $\times$
        & \textbf{4.56}
        & \textbf{3.57$\times$}
        & \textbf{1041.72}
        & \textbf{3.57$\times$}
        & 1.0469
        & 27.862
        & 17.529
        & 0.6838
        & 0.3103 \\

        $\times$ & $\checkmark$
        & 5.83
        & 2.79$\times$
        & 1117.22
        & 3.32$\times$
        & 1.0405
        & 27.951
        & 20.018
        & 0.7417
        & 0.2282 \\

        $\checkmark$ & $\checkmark$
        & 4.90
        & 3.32$\times$
        & 1116.12
        & 3.33$\times$
        & 1.0413
        & \textbf{27.971}
        & \textbf{20.115}
        & \textbf{0.7435}
        & \textbf{0.2279} \\

        \bottomrule
    \end{tabular}
\end{table}
\section{Conclusion}
We presented \textbf{TORE}, a training-free diffusion acceleration framework that selects token- and horizon-specific Taylor orders by estimating residual prediction errors from cached full-model residuals. The estimates account for truncation and finite-difference derivative errors without requiring future ground truth. Precomputed order maps enable skipped steps to forecast residuals while bypassing the entire denoiser. Experiments on text-to-image, text-to-video, and class-to-image generation demonstrate favorable quality--efficiency trade-offs and validate our error estimates and adaptive order selection.
\section*{AI use statement}
Generative AI tools were used solely for language editing and polishing, including improving grammar, wording, clarity, and readability of the manuscript. All AI-assisted edits were reviewed and revised by the authors. The authors take full responsibility for the final content of the paper.

\bibliography{iclr2027_conference}

@inproceedings{Song2021DDIM,
author = {Song, Jiaming and Meng, Chenlin and Ermon, Stefano},
booktitle = {International Conference on Learning Representations},
title = {Denoising Diffusion Implicit Models},
year = {2021}
}

@inproceedings{Rombach2022LDM,
author = {Rombach, Robin and Blattmann, Andreas and Lorenz, Dominik and Esser, Patrick and Ommer, Bj{\"o}rn},
booktitle = {Proceedings of the IEEE/CVF Conference on Computer Vision and Pattern Recognition},
pages = {10684--10695},
title = {High-Resolution Image Synthesis with Latent Diffusion Models},
year = {2022}
}

@inproceedings{Ho2022VideoDiffusion,
author = {Ho, Jonathan and Salimans, Tim and Gritsenko, Alexey and Chan, William and Norouzi, Mohammad and Fleet, David J.},
booktitle = {Advances in Neural Information Processing Systems},
volume = {35},
pages = {8633--8646},
title = {Video Diffusion Models},
year = {2022}
}

@inproceedings{peebles2023scalable,
  title     = {Scalable Diffusion Models with Transformers},
  author    = {Peebles, William and Xie, Saining},
  booktitle = {Proceedings of the IEEE/CVF International Conference on Computer Vision},
  pages     = {4195--4205},
  year      = {2023},
  doi       = {10.1109/ICCV51070.2023.00387}
}

@inproceedings{Zhu2026TAP,
author = {Zhu, Haowei and Huang, Tingxuan and Wang, Xing and Zhao, Tianyu and Wang, Jiexi and Chen, Weifeng and Peng, Xurui and Chen, Fangmin and Yong, Junhai and Wang, Bin},
booktitle = {Proceedings of the IEEE/CVF Conference on Computer Vision and Pattern Recognition},
pages = {36082--36091},
title = {{TAP}: A Token-Adaptive Predictor Framework for Training-Free Diffusion Acceleration},
year = {2026}
}

@inproceedings{Liu2022PNDM,
author = {Liu, Luping and Ren, Yi and Lin, Zhijie and Zhao, Zhou},
booktitle = {International Conference on Learning Representations},
title = {Pseudo Numerical Methods for Diffusion Models on Manifolds},
year = {2022}
}

@inproceedings{Lu2022DPMSolver,
author = {Lu, Cheng and Zhou, Yuhao and Bao, Fan and Chen, Jianfei and Li, Chongxuan and Zhu, Jun},
booktitle = {Advances in Neural Information Processing Systems},
pages = {5775--5787},
title = {{DPM-Solver}: A Fast {ODE} Solver for Diffusion Probabilistic Model Sampling in Around 10 Steps},
volume = {35},
year = {2022}
}

@inproceedings{Zhang2023DEIS,
author = {Zhang, Qinsheng and Chen, Yongxin},
booktitle = {International Conference on Learning Representations},
title = {Fast Sampling of Diffusion Models with Exponential Integrator},
year = {2023}
}

@inproceedings{Zhao2023UniPC,
author = {Zhao, Wenliang and Bai, Lujia and Rao, Yongming and Zhou, Jie and Lu, Jiwen},
booktitle = {Advances in Neural Information Processing Systems},
pages = {49842--49869},
title = {{UniPC}: A Unified Predictor-Corrector Framework for Fast Sampling of Diffusion Models},
volume = {36},
year = {2023}
}

@inproceedings{Salimans2022ProgressiveDistillation,
author = {Salimans, Tim and Ho, Jonathan},
booktitle = {International Conference on Learning Representations},
title = {Progressive Distillation for Fast Sampling of Diffusion Models},
year = {2022}
}

@inproceedings{Song2023ConsistencyModels,
author = {Song, Yang and Dhariwal, Prafulla and Chen, Mark and Sutskever, Ilya},
booktitle = {Proceedings of the 40th International Conference on Machine Learning},
pages = {32211--32252},
publisher = {PMLR},
series = {Proceedings of Machine Learning Research},
title = {Consistency Models},
volume = {202},
year = {2023}
}

@article{Luo2023LatentConsistency,
author  = {Luo, Simian and Tan, Yiqin and Huang, Longbo and Li, Jian and Zhao, Hang},
title   = {Latent Consistency Models: Synthesizing High-Resolution Images with Few-Step Inference},
journal = {arXiv preprint arXiv:2310.04378},
year    = {2023}
}

@inproceedings{Yin2024DMD,
author = {Yin, Tianwei and Gharbi, Micha{\"e}l and Zhang, Richard and Shechtman, Eli and Durand, Fr{\'e}do and Freeman, William T. and Park, Taesung},
booktitle = {Proceedings of the IEEE/CVF Conference on Computer Vision and Pattern Recognition},
pages = {6613--6623},
title = {One-Step Diffusion with Distribution Matching Distillation},
year = {2024}
}

@inproceedings{So2024FRDiff,
author = {So, Junhyuk and Lee, Jungwon and Park, Eunhyeok},
booktitle = {European Conference on Computer Vision},
title = {{FRDiff}: Feature Reuse for Universal Training-Free Acceleration of Diffusion Models},
year = {2024}
}

@inproceedings{Li2024FasterDiffusion,
author = {Li, Senmao and Hu, Taihang and van de Weijer, Joost and Khan, Fahad Shahbaz and Liu, Tao and Li, Linxuan and Yang, Shiqi and Wang, Yaxing and Cheng, Ming-Ming and Yang, Jian},
booktitle = {Advances in Neural Information Processing Systems},
title = {Faster Diffusion: Rethinking the Role of the Encoder for Diffusion Model Inference},
volume = {37},
year = {2024}
}

@inproceedings{Ma2024DeepCache,
author = {Ma, Xinyin and Fang, Gongfan and Wang, Xinchao},
booktitle = {Proceedings of the IEEE/CVF Conference on Computer Vision and Pattern Recognition},
pages = {15762--15772},
title = {{DeepCache}: Accelerating Diffusion Models for Free},
year = {2024}
}

@inproceedings{Wimbauer2024BlockCaching,
author = {Wimbauer, Felix and Wu, Bichen and Schoenfeld, Edgar and Dai, Xiaoliang and Hou, Ji and He, Zijian and Sanakoyeu, Artsiom and Zhang, Peizhao and Tsai, Sam and Kohler, Jonas and Rupprecht, Christian and Cremers, Daniel and Vajda, Peter and Wang, Jialiang},
booktitle = {Proceedings of the IEEE/CVF Conference on Computer Vision and Pattern Recognition},
pages = {6211--6220},
title = {Cache Me if You Can: Accelerating Diffusion Models through Block Caching},
year = {2024}
}

@article{Chen2024DeltaDiT,
author = {Chen, Pengtao and Shen, Mingzhu and Ye, Peng and Cao, Jianjian and Tu, Chongjun and Bouganis, Christos-Savvas and Zhao, Yiren and Chen, Tao},
journal = {arXiv preprint arXiv:2406.01125},
title = {{$\Delta$-DiT}: A Training-Free Acceleration Method Tailored for Diffusion Transformers},
year = {2024}
}

@article{Selvaraju2024FORA,
author = {Selvaraju, Pratheba and Ding, Tianyu and Chen, Tianyi and Zharkov, Ilya and Liang, Luming},
journal = {arXiv preprint arXiv:2407.01425},
title = {{FORA}: Fast-Forward Caching in Diffusion Transformer Acceleration},
year = {2024}
}

@inproceedings{Ma2024LearningToCache,
author = {Ma, Xinyin and Fang, Gongfan and Mi, Michael Bi and Wang, Xinchao},
booktitle = {Advances in Neural Information Processing Systems},
title = {Learning-to-Cache: Accelerating Diffusion Transformer via Layer Caching},
volume = {37},
year = {2024}
}

@inproceedings{Zhao2025PAB,
author = {Zhao, Xuanlei and Jin, Xiaolong and Wang, Kai and You, Yang},
booktitle = {International Conference on Learning Representations},
title = {Real-Time Video Generation with Pyramid Attention Broadcast},
year = {2025}
}

@inproceedings{Zou2025ToCa,
author = {Zou, Chang and Liu, Xuyang and Liu, Ting and Huang, Siteng and Zhang, Linfeng},
booktitle = {International Conference on Learning Representations},
title = {Accelerating Diffusion Transformers with Token-Wise Feature Caching},
year = {2025}
}

@inproceedings{Liu2025TeaCache,
author = {Liu, Feng and Zhang, Shiwei and Wang, Xiaofeng and Wei, Yujie and Qiu, Haonan and Zhao, Yuzhong and Zhang, Yingya and Ye, Qixiang and Wan, Fang},
booktitle = {Proceedings of the IEEE/CVF Conference on Computer Vision and Pattern Recognition},
pages = {7353--7363},
title = {Timestep Embedding Tells: It's Time to Cache for Video Diffusion Model},
year = {2025}
}

@inproceedings{Kahatapitiya2025AdaCache,
author = {Kahatapitiya, Kumara and Liu, Haozhe and He, Sen and Liu, Ding and Jia, Menglin and Zhang, Chenyang and Ryoo, Michael S. and Xie, Tian},
booktitle = {Proceedings of the IEEE/CVF International Conference on Computer Vision},
title = {Adaptive Caching for Faster Video Generation with Diffusion Transformers},
year = {2025}
}

@inproceedings{Zheng2025ClusterCache,
author = {Zheng, Zhixin and Wang, Xinyu and Zou, Chang and Wang, Shaobo and Zhang, Linfeng},
booktitle = {Proceedings of the 33rd ACM International Conference on Multimedia},
pages = {10181--10189},
publisher = {Association for Computing Machinery},
title = {Compute Only 16 Tokens in One Timestep: Accelerating Diffusion Transformers with Cluster-Driven Feature Caching},
year = {2025}
}

@inproceedings{Liu2025TaylorSeer,
author = {Liu, Jiacheng and Zou, Chang and Lyu, Yuanhuiyi and Chen, Junjie and Zhang, Linfeng},
booktitle = {Proceedings of the IEEE/CVF International Conference on Computer Vision},
pages = {15853--15863},
title = {From Reusing to Forecasting: Accelerating Diffusion Models with {TaylorSeers}},
year = {2025}
}

@article{Guan2025ConfidenceGatedTaylor,
author = {Guan, Xiaoliu and Jiang, Lielin and Chen, Hanqi and Zhang, Xu
        and Yan, Jiaxing and Wang, Guanzhong and Liu, Yi
        and Zhang, Zetao and Wu, Yu},
title = {Forecasting When to Forecast: Accelerating Diffusion Models with Confidence-Gated Taylor},
journal = {Knowledge-Based Systems},
volume = {330},
pages = {114635},
year = {2025},
doi = {10.1016/j.knosys.2025.114635}
}

@article{Cui2026PredictToSkip,
author = {Cui, Hanshuai and Tang, Zhiqing and Ma, Qianli and Yao, Zhi and Jia, Weijia},
journal = {arXiv preprint arXiv:2602.18093},
title = {Predict to Skip: Linear Multistep Feature Forecasting for Efficient Diffusion Transformers},
year = {2026}
}

@inproceedings{feng2026hicache,
  title={HiCache: A plug-in scaled-Hermite upgrade for Taylor-style cache-then-forecast diffusion acceleration},
  author={Feng, Liang and Zheng, Shikang and Liu, Jiacheng and Lin, Yuqi and Zhou, Qinming and Cai, Peiliang and Wang, Xinyu and Chen, Junjie and Zou, Chang and Ma, Yue and others},
  booktitle={International Conference on Learning Representations},
  volume={2026},
  pages={112361--112386},
  year={2026}
}

@inproceedings{Han2026Spectrum,
author = {Han, Jiaqi and Shi, Juntong and Li, Puheng and Ye, Haotian and Guo, Qiushan and Ermon, Stefano},
booktitle = {Proceedings of the IEEE/CVF Conference on Computer Vision and Pattern Recognition},
title = {Adaptive Spectral Feature Forecasting for Diffusion Sampling Acceleration},
year = {2026}
}

@inproceedings{Shen2026L2P,
author = {Shen, Zhirong and Huang, Rui and Liu, Jiacheng and Zou, Chang and Cai, Peiliang and Zheng, Shikang and Shi, Zhengyi and Feng, Liang and Zhang, Linfeng},
booktitle = {Proceedings of the IEEE/CVF Conference on Computer Vision and Pattern Recognition},
title = {Beyond Fixed Formulas: Data-Driven Linear Predictor for Efficient Diffusion Models},
year = {2026}
}

@inproceedings{He2026TCPade,
author = {He, Shaoxuan and Cui, Benlei and Huang, Bukun and Ye, Zhizeng
        and Sun, Yunyun and Huang, Longtao and Xue, Hui and Yang, Yang
        and Hong, Haiwen and Tang, Jingqun and Zhao, Zhou},
title = {{TC-Pad\'e}: Trajectory-Consistent Pade Approximation for Diffusion Acceleration},
booktitle = {Proceedings of the IEEE/CVF Conference on Computer Vision and Pattern Recognition},
pages = {35768--35778},
year = {2026}
}

@article{labs2024flux,
  title={{Flux}},
  author={{Black Forest Labs}},
  journal={https://github.com/black-forest-labs/flux},
  year={2024}
}

@article{wan2025wan,
  title={Wan: Open and advanced large-scale video generative models},
  author={Wan, Team and Wang, Ang and Ai, Baole and Wen, Bin and Mao, Chaojie and Xie, Chen-Wei and Chen, Di and Yu, Feiwu and Zhao, Haiming and Yang, Jianxiao and others},
  journal={arXiv preprint arXiv:2503.20314},
  year={2025}
}
\bibliographystyle{iclr2027_conference}

\appendix
\section{Algorithm}
\label{app:Algorithm}
Algorithm~\ref{alg:adaptive_prediction} summarizes the inference procedure of TORE. At each full-computation step, TORE updates the residual cache and preselects a Taylor order for each token at each of the next $N-1$ skip steps. Skip steps then retrieve the selected orders and reconstruct the network output from the input and the predicted residual.

\begin{algorithm}[H]
\caption{Token-wise Error-Guided Residual Forecasting}
\label{alg:adaptive_prediction}
\small
\KwRequire{sampling steps $T$, full-step interval $N$,
maximum order $M$}

Initialize residual cache $\mathcal{C}$ and order map $\mathcal{O}$\;

\For{$\tau = T, T-1, \ldots, 1$}{
    \eIf{$\tau$ is a full-computation step}{
        $t \leftarrow \tau$\;

        $x_{\mathrm{out}}(t)
        \leftarrow
        \Phi_{\theta}(x_{\mathrm{in}}(t), t)$\;

        $R(t)
        \leftarrow
        x_{\mathrm{out}}(t)-x_{\mathrm{in}}(t)$\;

        Update $\mathcal{C}$ with $\{R_l(t)\}_l$
        and the corresponding finite differences\;

        $\mathcal{O} \leftarrow \varnothing$\;

        \For{$k = 1$ \KwTo $N-1$}{
            \ForEach{token $l$}{
                Estimate
                $\{\widehat{E}_{l,m}(t-k)\}_{m=0}^{M}$
                from $\mathcal{C}$ using
                Eqs.~(\ref{eq:truncation_error})--(\ref{eq:fd_error})\;

                $\widehat{m}_l(t-k)
                \leftarrow
                \arg\min_{m\in\{0,\ldots,M\}}
                \widehat{E}_{l,m}(t-k)$\;

                $\mathcal{O}[k,l]
                \leftarrow
                \widehat{m}_l(t-k)$\;
            }
        }
    }{
        $k \leftarrow t-\tau$\;

        \ForEach{token $l$}{
            $\widehat{m}_l(t-k)
            \leftarrow
            \mathcal{O}[k,l]$\;

            $\widehat{R}_{l,\widehat{m}_l(t-k)}(t-k)
            \leftarrow
            \mathrm{TaylorPredict}
            \bigl(
                \mathcal{C}_l,
                \widehat{m}_l(t-k),
                k
            \bigr)$\;
        }

        $\widehat{x}_{\mathrm{out}}(t-k)
        \leftarrow
        x_{\mathrm{in}}(t-k)
        +
        \widehat{R}(t-k)$\;
    }
}
\end{algorithm}
\section{Derivation of the Token-wise Error Estimator}
\label{app:error_derivation}
Let $R_l(t)$ denote the full-model residual of the $l$-th token at the current full-computation step $t$. Consecutive full-computation steps are separated by an interval $N$. Since denoising proceeds toward smaller timesteps, $t+N,t+2N,\ldots$ correspond to previously evaluated full steps whose residuals are available in the cache. We define the forward finite difference over the full-step interval $N$ as
\begin{equation}
    \Delta_N R_l(t)
    =
    R_l(t+N)-R_l(t),
\end{equation}
and recursively define the higher-order differences as
\begin{equation}
    \Delta_N^i R_l(t)
    =
    \Delta_N^{i-1}R_l(t+N)
    -
    \Delta_N^{i-1}R_l(t),
    \qquad i\geq 2.
\end{equation}

For a skip step $t-k$, where $k\in\{1,\ldots,N-1\}$ denotes the prediction horizon from the current full step, the $m$-th order Taylor prediction is
\begin{equation}
\label{eq:app_predictor}
    \widehat{R}_{l,m}(t-k)
    =
    R_l(t)
    +
    \sum_{i=1}^{m}
    \frac{(-k)^i}{i!N^i}
    \Delta_N^i R_l(t).
\end{equation}

\paragraph{Error decomposition.}
Expanding the target residual $R_l(t-k)$ around the current full step $t$ gives
\begin{equation}
    R_l(t-k)
    =
    R_l(t)
    +
    \sum_{i=1}^{m}
    \frac{(-k)^i}{i!}
    R_l^{(i)}(t)
    +
    \mathcal{T}_{l,m+1}(t,k),
\end{equation}
where $\mathcal{T}_{l,m+1}(t,k)$ denotes the Taylor truncation remainder after order $m$. Defining the prediction error as
\begin{equation}
    e_{l,m}(t-k)
    =
    R_l(t-k)
    -
    \widehat{R}_{l,m}(t-k),
\end{equation}
we obtain
\begin{equation}
\label{eq:app_error_decomposition}
\begin{aligned}
    e_{l,m}(t-k)
    =
    \mathcal{T}_{l,m+1}(t,k)
    +
    \sum_{i=1}^{m}
    \frac{(-k)^i}{i!}
    \left(
        R_l^{(i)}(t)
        -
        \frac{\Delta_N^iR_l(t)}{N^i}
    \right).
\end{aligned}
\end{equation}
The prediction error therefore arises from two sources: the Taylor truncation error and the error introduced by approximating the derivatives with finite differences.

\paragraph{Finite-difference error estimation.}
For the $i$-th order forward finite difference, a local expansion around $t$ gives
\begin{equation}
\label{eq:app_fd_expansion}
    \frac{\Delta_N^iR_l(t)}{N^i}
    =
    R_l^{(i)}(t)
    +
    \frac{iN}{2}R_l^{(i+1)}(t)
    +
    \mathcal{O}(N^2).
\end{equation}
The adjacent finite difference similarly satisfies
\begin{equation}
    \frac{\Delta_N^{i+1}R_l(t)}{N^{i+1}}
    =
    R_l^{(i+1)}(t)
    +
    \mathcal{O}(N).
\end{equation}
Substituting the latter into ~\eqref{eq:app_fd_expansion} eliminates the unknown $(i+1)$-th derivative and yields
\begin{equation}
\label{eq:app_fd_proxy}
    R_l^{(i)}(t)
    -
    \frac{\Delta_N^iR_l(t)}{N^i}
    =
    -
    \frac{i}{2N^i}
    \Delta_N^{i+1}R_l(t)
    +
    \mathcal{O}(N^2).
\end{equation}
Thus, although the original finite-difference approximation error contains a first-order term in $N$, the residual error of this approximation after substituting the adjacent finite difference starts from second order.

For convenience, we define the magnitude of the $i$-th order cached finite difference as
\begin{equation}
\label{eq:app_difference_magnitude}
    d_{l,i}(t)
    =
    \left\|
        \Delta_N^iR_l(t)
    \right\|_2.
\end{equation}
Using ~\eqref{eq:app_fd_proxy}, the finite-difference derivative error associated with an $m$-th order predictor is estimated by
\begin{equation}
\label{eq:app_fd_error}
    \widehat{E}_{l,m}^{\mathrm{FD}}(t-k)
    =
    \frac{1}{2}
    \sum_{i=1}^{m}
    \frac{k^i}{(i-1)!N^i}
    d_{l,i+1}(t).
\end{equation}

\paragraph{Truncation error estimation.}
The Taylor truncation error is approximated by the first neglected term, whose $(m+1)$-th derivative is in turn approximated using the corresponding cached finite difference:
\begin{equation}
\label{eq:app_trunc_error}
    \widehat{E}_{l,m}^{\mathrm{trunc}}(t-k)
    =
    \frac{k^{m+1}}
    {(m+1)!N^{m+1}}
    d_{l,m+1}(t).
\end{equation}

Combining the two error sources gives the computable residual prediction error surrogate
\begin{equation}
\label{eq:app_error_estimator}
\begin{aligned}
    \widehat{E}_{l,m}(t-k)
    &=
    \widehat{E}_{l,m}^{\mathrm{trunc}}(t-k)
    +
    \widehat{E}_{l,m}^{\mathrm{FD}}(t-k)
    \\
    &=
    \frac{k^{m+1}}
    {(m+1)!N^{m+1}}
    d_{l,m+1}(t)
    +
    \frac{1}{2}
    \sum_{i=1}^{m}
    \frac{k^i}
    {(i-1)!N^i}
    d_{l,i+1}(t).
\end{aligned}
\end{equation}

\paragraph{Token-wise adaptive order selection.}
Since different tokens exhibit different residual dynamics, TORE selects the prediction order independently for each token and each prediction horizon:
\begin{equation}
\label{eq:app_order_selection}
    \widehat{m}_l(t-k)
    =
    \underset{m\in\{0,\ldots,M\}}{\arg\min}
    \widehat{E}_{l,m}(t-k),
    \qquad
    k\in\{1,\ldots,N-1\}.
\end{equation}
For $m=0$, the finite-difference derivative term vanishes and the estimator reduces to
\begin{equation}
    \widehat{E}_{l,0}(t-k)
    =
    \frac{k}{N}
    d_{l,1}(t),
\end{equation}
corresponding to direct reuse of the residual at the current full step, i.e., $\widehat{R}_{l,0}(t-k)=R_l(t)$.

For further interpretation, we define the estimated error reduction when increasing the prediction order from $m-1$ to $m$ as
\begin{equation}
    \widehat{G}_{l,m}(t-k)
    =
    \widehat{E}_{l,m-1}(t-k)
    -
    \widehat{E}_{l,m}(t-k),
    \qquad m\geq 1.
\end{equation}
Substituting ~\eqref{eq:app_error_estimator} gives
\begin{equation}
\label{eq:app_gain}
    \widehat{G}_{l,m}(t-k)
    =
    \frac{k^m}{m!N^m}
    \left[
        d_{l,m}(t)
        -
        \left(
            \frac{k}{(m+1)N}
            +
            \frac{m}{2}
        \right)
        d_{l,m+1}(t)
    \right].
\end{equation}
Therefore, increasing the prediction order from $m-1$ to $m$ is estimated to be beneficial when $\widehat{G}_{l,m}(t-k)>0$, i.e., when the variation captured at the current order is sufficiently large relative to the remaining higher-order variation.

\section{Additional Experimental Results}
\subsection{Analysis of Error Components}

\begin{figure}[!ht]
    \centering
    \includegraphics[width=0.55\linewidth]{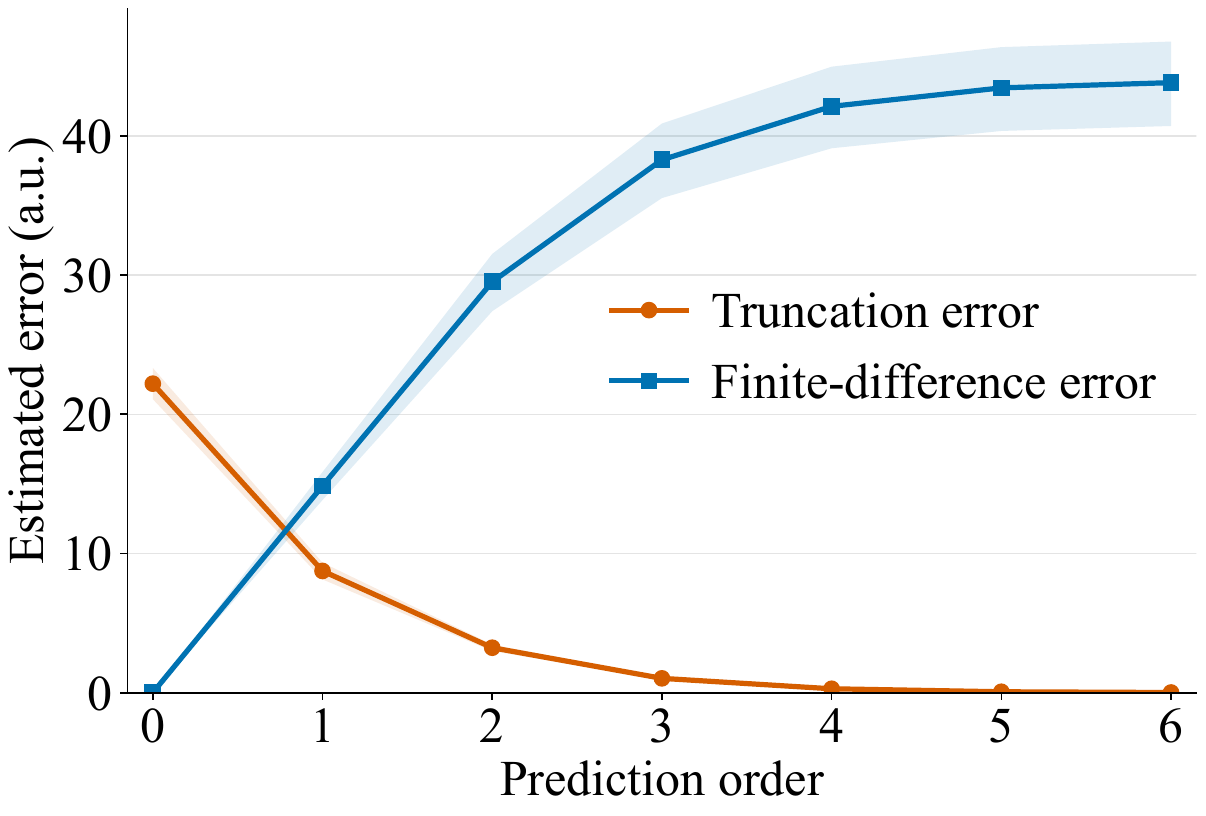}
    \caption{Variation of estimated truncation and finite-difference approximation errors across Taylor orders. The two components exhibit opposing trends, motivating their joint consideration in adaptive order selection.}
    \label{fig:app_error}
\end{figure}

To further examine the roles of the two error components, we analyze their estimated magnitudes across different Taylor orders. As shown in Figure~\ref{fig:app_error}, the truncation error generally decreases with increasing order, while the finite-difference approximation error increases. These opposing trends motivate combining both estimates to balance the benefits and approximation costs of higher-order forecasting.

\subsection{Quality--Cost Trade-off with Fixed-Order Predictors}

\begin{figure}[t]
    \centering
    \includegraphics[width=0.65\linewidth]{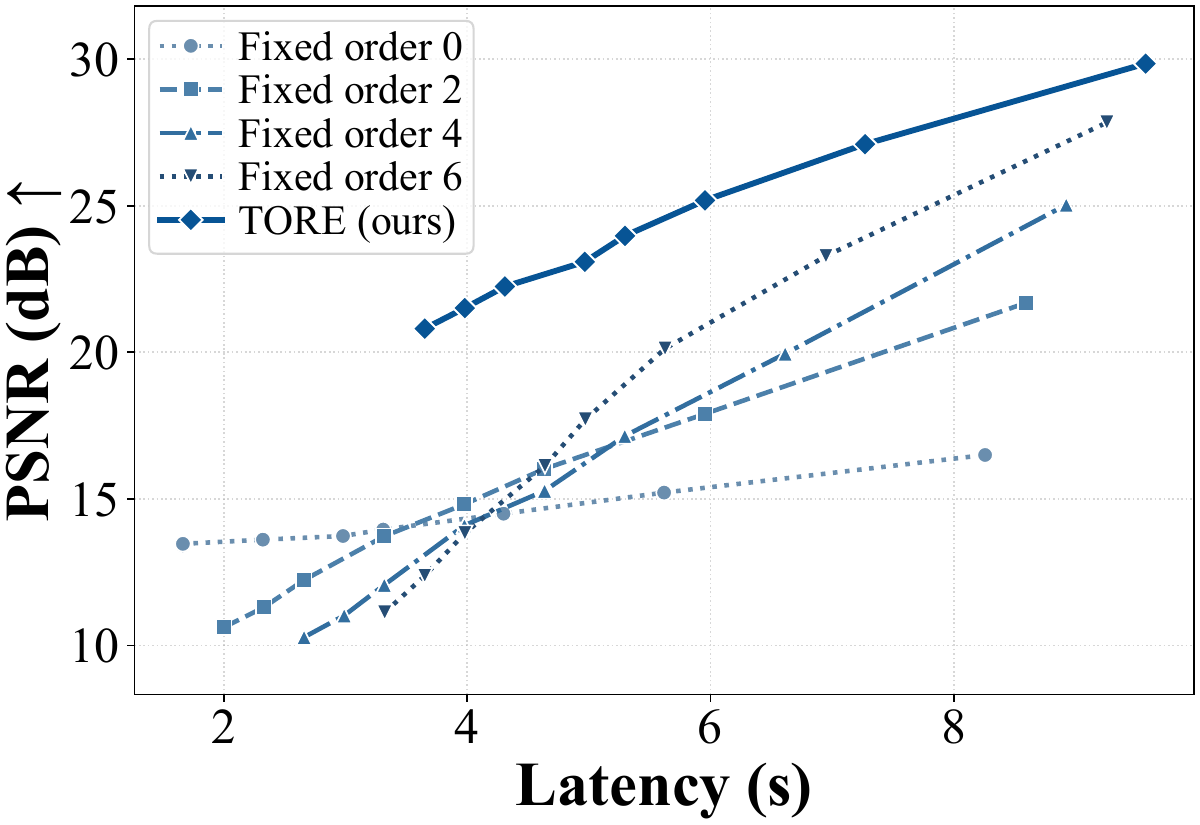}
    \caption{Quality--latency trade-off between TORE and fixed-order Taylor predictors. Different computational budgets are obtained by varying the full-computation interval $N$. All reported latencies include warm-up computation.}
    \label{fig:app_costtrade}
\end{figure}

We further compare TORE with fixed-order Taylor predictors across different computational budgets. We vary the full-computation interval $N$ to obtain different latency levels. For this controlled comparison, each method performs sufficient initial full-model evaluations before forecasting, with all warm-up computation included in the reported latency. As shown in Figure~\ref{fig:app_costtrade}, TORE achieves a more favorable PSNR--latency trade-off than the evaluated fixed-order predictors.

\subsection{Sensitivity to Maximum Taylor Order}
\begin{figure}[t]
    \centering
    \includegraphics[width=0.8\linewidth]{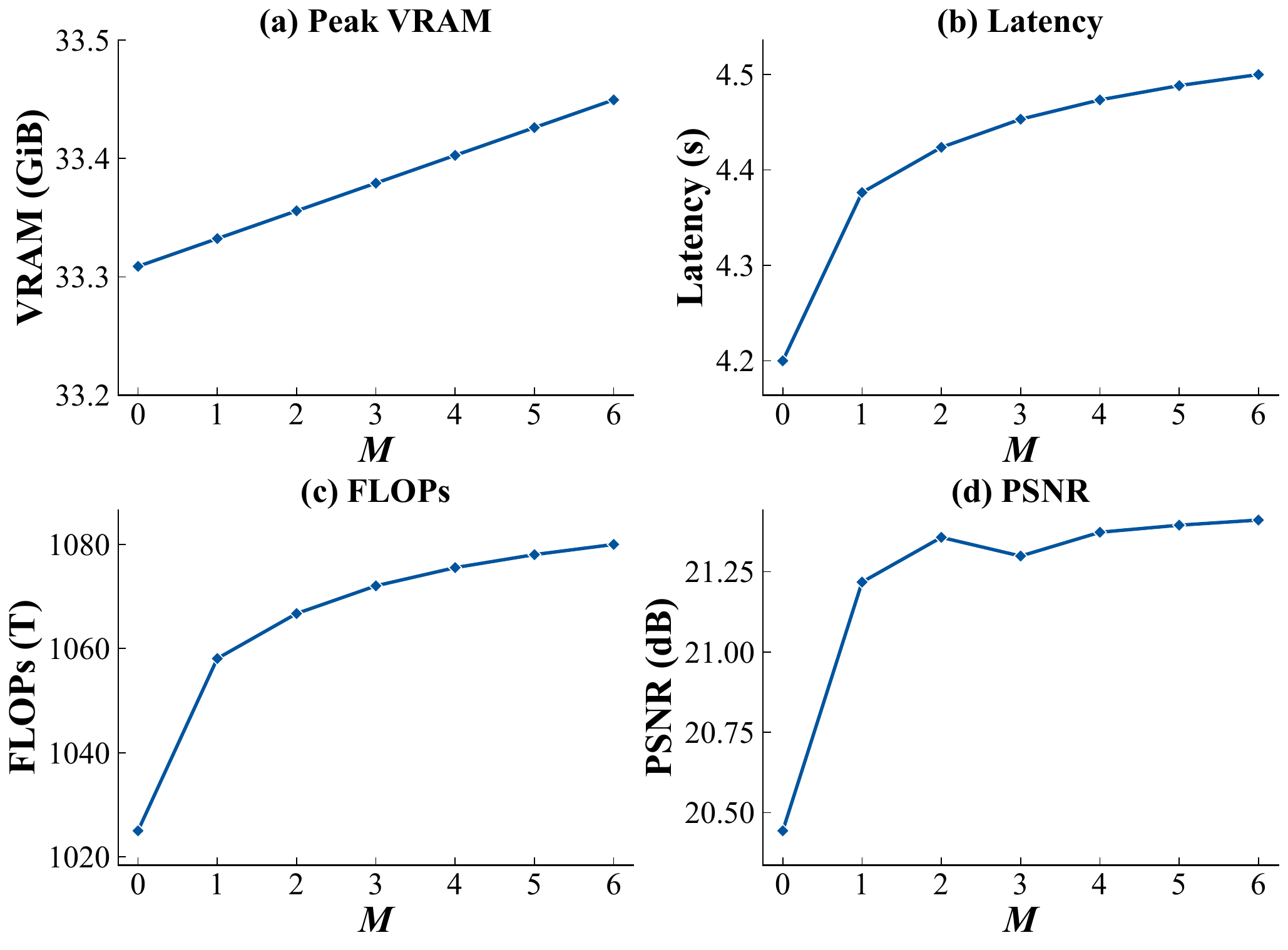}
    \caption{Sensitivity to the maximum Taylor order $M$ on FLUX.1-dev. (a) Peak VRAM, (b) inference latency, (c) FLOPs, and (d) PSNR.}
    \label{fig:app_m_sensitivity}
\end{figure}

We further investigate the effects of the maximum Taylor order $M$ on generation fidelity and inference overhead. Experiments are conducted on FLUX.1-dev using a single NVIDIA RTX PRO 6000 GPU, with $N=7$. We vary $M$ from 0 to 6 and use the same random seed across all configurations. As shown in Fig.~\ref{fig:app_m_sensitivity}, increasing $M$ improves PSNR overall, with diminishing gains at higher orders. Meanwhile, latency, FLOPs, and peak VRAM increase moderately. These results illustrate the trade-off between generation fidelity and inference overhead.

\subsection{Distribution of Selected Taylor Orders}
To examine the actual utilization of different Taylor orders, we randomly select 11 prompts and analyze the selected-order distributions for three main-table configurations: $(N,M)=(3,4)$, $(5,6)$, and $(8,5)$. We divide the 50 denoising steps into early (1--17), middle (18--34), and late (35--50) stages.

\begin{figure}[t]
    \centering
    \includegraphics[width=\linewidth]{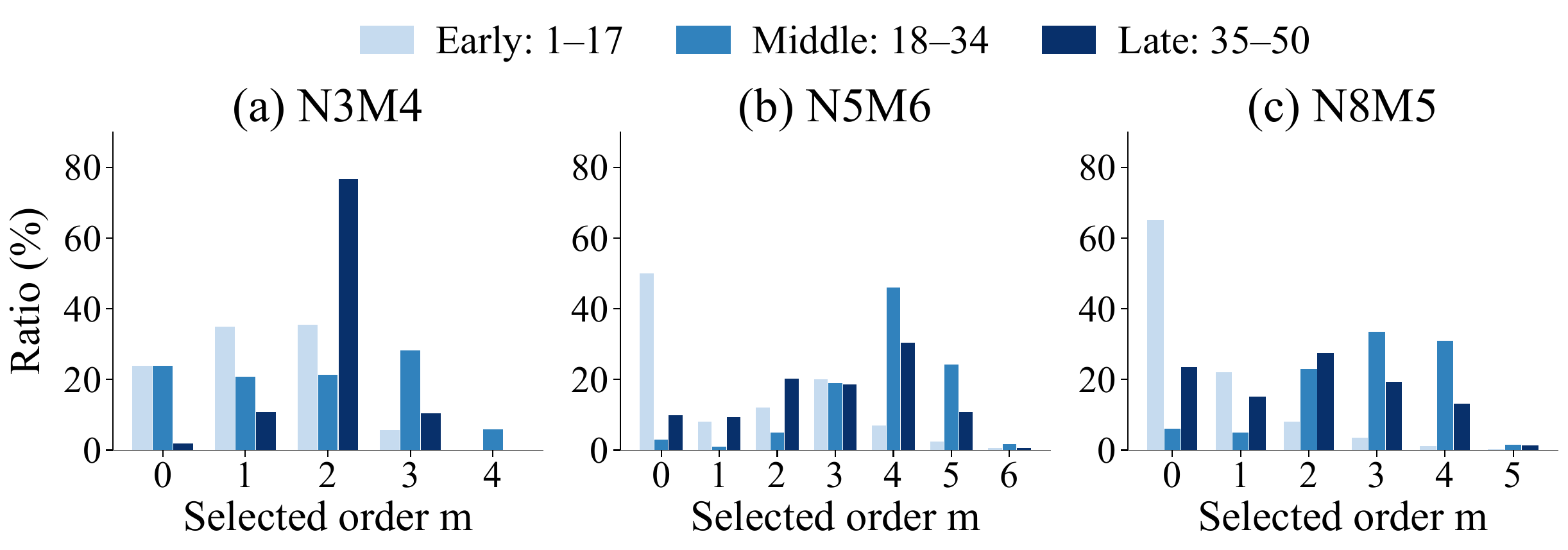}
    \caption{Stage-wise distribution of selected Taylor orders for three main-table configurations: (a) $N=3,M=4$, (b) $N=5,M=6$, and (c) $N=8,M=5$. Results are collected from 11 randomly selected prompts and grouped into early (steps 1--17), middle (18--34), and late (35--50) denoising stages. Ratios are computed over the selected orders of skip-step tokens within each stage.}
    \label{fig:app_order_distribution}
\end{figure}

As shown in Fig.~\ref{fig:app_order_distribution}, the selected orders exhibit distinct stage-dependent patterns. Lower orders are more frequently selected in the early stage, partly due to limited residual history. Higher-order selections become more frequent in the middle stage, while the late stage generally favors intermediate orders. 

\subsection{Qualitative Results across Acceleration Levels}
\begin{figure}[!ht]
    \centering
    \includegraphics[width=0.75\linewidth]{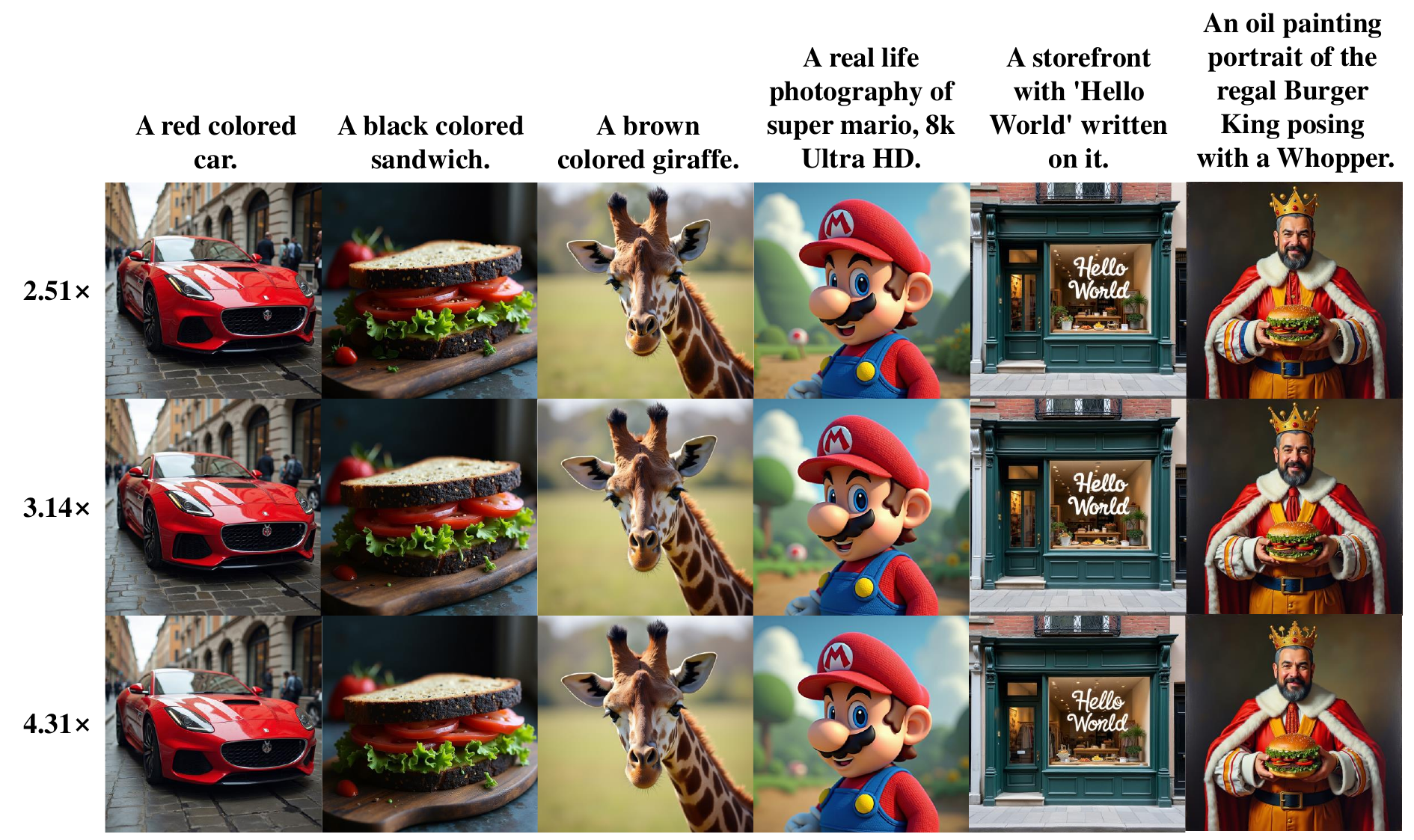}
    \caption{
    Qualitative results of TORE under increasing acceleration.
    We show text-to-image generation results at $2.51\times$, $3.14\times$,
    and $4.31\times$ latency speedups. TORE preserves semantic content and
    overall visual quality as the acceleration becomes more aggressive,
    demonstrating its robustness across different acceleration levels.
    }
    \label{fig:additional_t2i}
\end{figure}

Fig.~\ref{fig:additional_t2i} further examines the behavior of TORE as the acceleration level increases. We compare generations obtained at $2.51\times$, $3.14\times$, and $4.31\times$ latency speedups using the same set of prompts. As the forecasting becomes more aggressive,
the generated samples largely preserve their semantic content and global visual structure, indicating that TORE remains stable across a broad range of acceleration levels.

\end{document}